\pdfoutput = 1

\documentclass[Afour,sageh,times]{sagej}
\usepackage{moreverb,url}
\usepackage[colorlinks,bookmarksopen,bookmarksnumbered,citecolor=red,urlcolor=red]{hyperref}
\usepackage{amsmath} 
\usepackage{amsmath,amssymb,amsfonts}
\usepackage{bm}
\usepackage{multirow}

\usepackage{endnotes}
\usepackage{paralist}
\usepackage{array}
\usepackage{tabularx} 
\usepackage{booktabs}

\newcommand\BibTeX{{\rmfamily B\kern-.05em \textsc{i\kern-.025em b}\kern-.08em
T\kern-.1667em\lower.7ex\hbox{E}\kern-.125emX}}

\def\volumeyear{2024}
  
\begin{document}

\runninghead{Kim et al.}

\title{Contact-Implicit Model Predictive Control:\\Controlling Diverse Quadruped Motions\\ Without Pre-Planned Contact Modes or Trajectories}
\author{Gijeong Kim\affilnum{1}, Dongyun Kang\affilnum{1}, Joon-Ha Kim\affilnum{1}, Seungwoo Hong\affilnum{2} and Hae-Won Park\affilnum{1}
}

\affiliation{\affilnum{1} Department of Mechanical Engineering, Korea Advanced Institute of Science and Technology, South Korea\\
\affilnum{2} Department of Mechanical Engineering, Massachusetts Institute of Technology, USA}
\corrauth{Hae-Won Park,  
Korea Advanced Institute of Science and Technology, 291 Daehak-ro, Yuseong-gu, 34141 Daejeon, South Korea.
}
\email{haewonpark@kaist.ac.kr}

\begin{abstract}
This paper presents a contact-implicit model predictive control (MPC) framework for the real-time discovery of multi-contact motions, without predefined contact mode sequences or foothold positions. 
This approach utilizes the contact-implicit differential dynamic programming (DDP) framework, merging the hard contact model with a linear complementarity constraint. 
We propose the analytical gradient of the contact impulse based on relaxed complementarity constraints to further the exploration of a variety of contact modes. 
By leveraging a hard contact model-based simulation and computation of search direction through a smooth gradient, our methodology identifies dynamically feasible state trajectories, control inputs, and contact forces while simultaneously unveiling new contact mode sequences.
However, the broadened scope of contact modes does not always ensure real-world applicability.
Recognizing this, we implemented differentiable cost terms to guide foot trajectories and make gait patterns. 
Furthermore, to address the challenge of unstable initial roll-outs in an MPC setting, we employ the multiple shooting variant of DDP. 
The efficacy of the proposed framework is validated through simulations and real-world demonstrations using a 45 kg HOUND quadruped robot, performing various tasks in simulation and showcasing actual experiments involving a forward trot and a front-leg rearing motion.
\end{abstract}


\maketitle
\section{Introduction}
Model Predictive Control (MPC)~\citep{cheetah3,Real-timeMPC,mastalli2023inverse,hong2020real} recently has become one of the most widely used methods in controlling legged robot systems in real experimental environment. With application to many legged-robot platforms, MPC demonstrates impressive performance, including high-speed running~\citep{Donghyun}, robust walking~\citep{Bledt}, perceptive locomotion~\citep{corberes2023perceptive,grandia2023perceptive}, push recovery~\citep{chen2023quadruped}, and complex dynamic motions~\citep{BDI}. 

\begin{figure}
    \includegraphics[width=1.0\columnwidth]{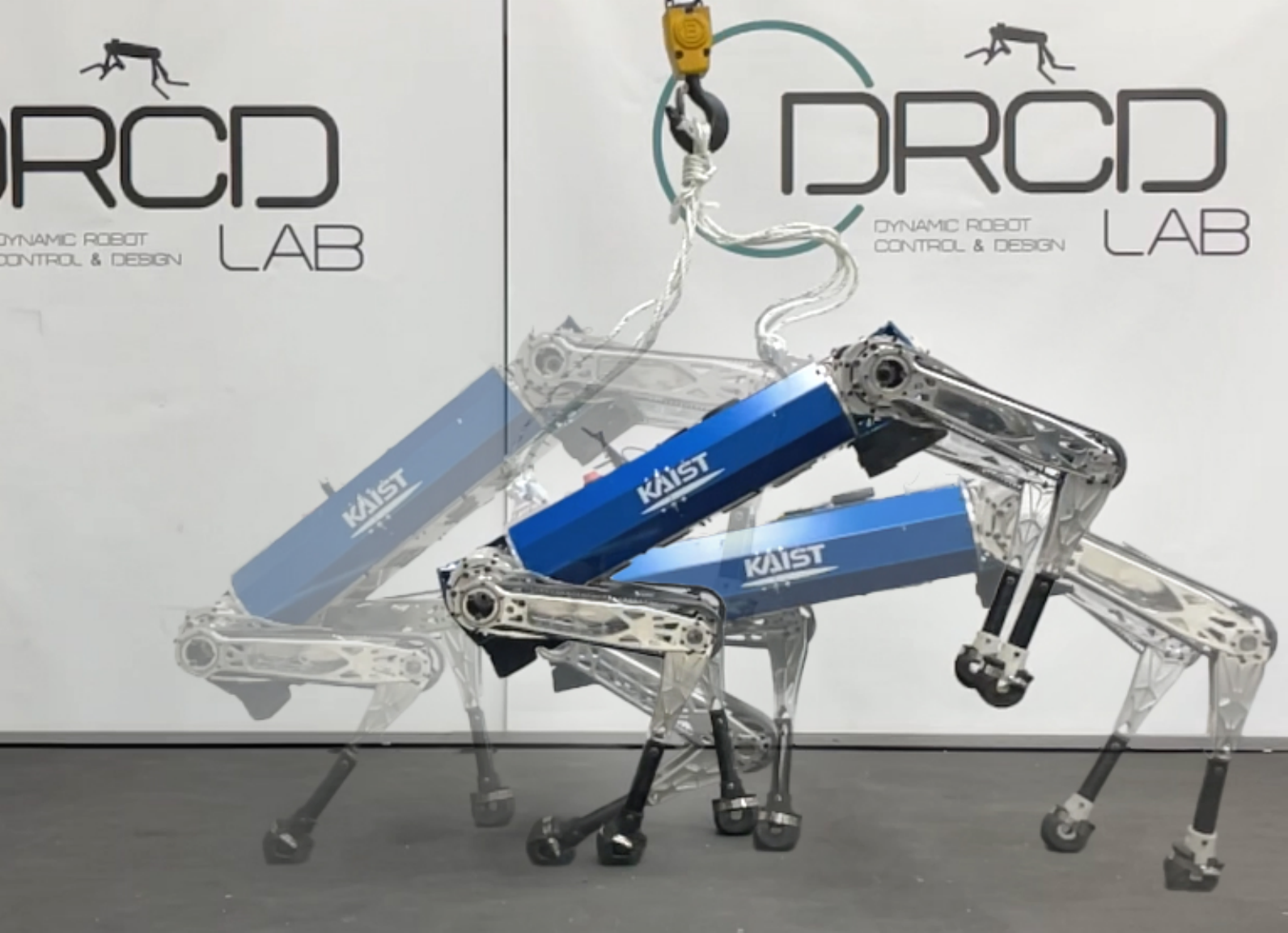}
    \caption[Experiment result for desired pitch angle reference configuration.]{Experimental demonstration of front-leg rearing motion with the HOUND quadruped robot using the proposed framework. Only a desired body pitch angle was set, without any predefined contact modes.}
    \label{figure:highlight3}
\end{figure}

However, previous approaches based on MPC focus on the choice of ground reaction forces or torque trajectory under the fixed contact modes, and only a few approaches consider the contact sequence~\citep{Bledt,neunert2018whole}. For dynamic maneuvers demanding non-repetitive and asymmetric contacts, prior formulations require presetting the \textit{feasible} contact modes before optimization. However, tackling this is complex due to the intertwined nature of overall motion patterns (e.g., body velocity, swing leg trajectory, and foothold position) with specific contact modes. Moreover, this predefined process for the entire horizon should be repeated for each distinct task.

To address the aforementioned challenges, this work employs a contact-implicit approach. This approach facilitates the concurrent discovery of \textit{feasible} contact modes, ensuring compatibility with states, control inputs, and contact forces. Based on the direct transcription method~\citep{Posa1,Manchester2}, the complementarity constraints resulting from the Signorini condition are incorporated into explicit constraints forms to ensure the contact feasibility. However, this explicit constraint can lead to computational burdens due to exponential contact mode combinations. To circumvent this issue, this work adopts a contact-implicit approach based on the differential dynamic programming (DDP) method~\citep{Yuval2012,TO-w-contact,carius2018trajectory}. The complementarity constraints are considered implicitly within the optimization process, mitigating computational burdens associated with the exponential growth of mode sequences as the number of contact points increases~\citep{carius2018trajectory}. Moreover, the utilization of DDP, exhibiting a linear increase in computation over the horizon, enables real-time computation for contact-implicit model predictive control.

Incorporating the hard contact model into the DDP framework enables the discovery of new contact modes. However, challenges arise due to the strict linear complementarity constraint, which tends to be stuck at the initial contact condition \citep{KarenLiu}. Specifically, in the contact case, the gradient directs towards maintaining zero contact velocity, thereby preventing contact breakage. To tackle this problem, we propose an analytical gradient based on relaxed complementarity constraints. This smooth gradient facilitates the discovery of new contact modes, leading to lower-cost solutions outperforming the non-relaxed cases.

While exploring diverse contact modes enriches motion repertoire, it does not ensure real-world robot implementation. Absent explicit foot trajectory considerations, motions often manifest insufficient foot clearance, resulting in issues such as minimal foot-lifting~\citep{Posa1} or prevalent scuffing~\citep{carius2019trajectory}. Additionally, a broader search space across contact modes might produce undesired results, such as three-leg walking of a quadruped robot. To refine these motions for practical use, we introduce differentiable cost terms to enhance foot clearance, minimize foot slippage, and produce gait patterns without periodic cost inputs. 

Lastly, leveraging the MPC scheme within the proposed framework enables the simultaneous generation and execution of multi-contact motions. However, the hybrid dynamics of the hard contact model lead to unstable initialization in the single shooting method, when the initialization relies solely on the control trajectory of the prior solution. This hybrid dynamics can induce numerous deviations in contact modes throughout the motion plan, even from slight initial state discrepancies~\citep{Posa1}. To address this, we utilize the multiple shooting variant of DDP. This ensures a consistent motion plan despite initial state discrepancies by enhancing robustness with state-control trajectory initialization.


The main contributions of the paper are as follows:
\begin{itemize}
\setlength\itemsep{0.05pt} 
\item A novel analytical gradient for contact impulses, based on relaxed complementarity constraint, is proposed for promoting the discovery of new contact modes.
\item For real-world implementations, we introduce specific details, including the integration of additional costs to ensure reliable motion and the use of the multiple shooting variant of DDP to enhance robustness.
\item The proposed contact-implicit MPC framework is demonstrated through 3D simulations of dynamic quadruped motions and real-world experiments on the 45kg \textit{KAIST HOUND} robot \citep{2022_Shin_HOUND}, including a front-leg rearing motion as shown in Figure~\ref{figure:highlight3} (also, Extension 1).
\end{itemize}

The remainder of this paper is structured as follows: Section~\ref{sec:related} discusses related work on trajectory optimization in contact-implicit approaches. Section~\ref{sec:background} provides background for our framework, and Section~\ref{sec:analytical_gradient_based_on_relaxation} introduces a novel analytical gradient of contact impulses in relaxed complementarity constraints. Section~\ref{sec:motion_generation} provides the entire structure of the optimal control problem, while Section~\ref{sec:implementation_motion_execution} offers implementation details of the proposed contact-implicit MPC. Finally, Section~\ref{sec:results} presents simulation and experimental results, followed by discussions (Section~\ref{sec:discussion}) and a summary (Section~\ref{sec:conclusion}).

In our previous work~\citep{kim2022contact}, we introduced a preliminary version of the contact-implicit DDP framework, restricted to 2D environments and focused on motion generation in idealized conditions. The novelty of this work is an expansion the framework to 3D environments for real-time execution, addressing robustness challenges within the contact-implicit MPC. We analyze the effects of relaxation variables within the complementarity constraint and present new 3D simulation results and real-world experiments on a quadruped robot with onboard computations.

\section{Related Work}
\label{sec:related}

\subsection{Contact-implicit Approach with Direct Transcription Method}
The contact-implicit approach with direct transcription method has been studied~\citep{Posa1,Manchester2,patel,moura2022non}. The contact impulse is optimized alongside state and control inputs, serving as an optimization variable. Therefore, the complementarity constraints are employed to model the Signorini condition and Coulomb friction, enforcing physically feasible contact motion. This approach formulates the problem as a mathematical program with complementarity constraints (MPCC), known for its challenging ill-posedness of the constraints. Consequently, relaxation methods \citep{Manchester2,patel} and penalty approaches \citep{mordatch2012contact,mordatch2012discovery,kurtz2023inverse} could be employed to make these constraints more tractable for convergence. However, applying relaxation or penalty methods may lead to infeasible contact motions, such as allowing contact forces to act at a non-zero distance, necessitating gradual tightening of relaxation variables for feasible motions. While these approaches generate various multi-contact motions without a priori contact mode sequences, their computational complexity and poor convergence often lead to their use for offline motion generations. 

Currently, progress for the real-time model predictive control has been made in~\citet{Manchester3,aydinoglu2022real,aydinoglu2023consensus}. \citet{aydinoglu2022real,aydinoglu2023consensus} tackle the MPCC problem using the alternating direction method of multipliers (ADMM), where the linear complementarity problem (LCP) constraints are enforced in the sub-problems within the ADMM iterations. Within these sub-problems, they handle smaller-scale mixed-integer quadratic programs (MIQP). This approach significantly reduces the complexity of the combinatorial search problems as compared to solving a singular large MIQP for the MPCC problem about the entire horizon. This approach enables real-time MPC for multi-contact systems, such as manipulation arms interacting with multiple rigid spheres, without a priori information about contact. 

\citet{Manchester3} enforce the contact dynamics using a time-varying LCP approximated about the reference trajectory. By utilizing a pre-computation and structure-exploiting interior point solver for this LCP, they could achieve real-time computation for contact-implicit MPC. By employing an interior-point method to solve the LCP, they compute smooth gradients using the central path parameter, relaxing the complementarity constraints and providing information about other contact modes. While this approach is capable of identifying new contact modes online, it relies on a reference trajectory computed offline in a contact-implicit manner~\citep{Manchester2}. This dependency requires distinct reference trajectories for different tasks, ultimately restricting the range of achievable motion types to the capabilities of offline trajectory optimization.

\subsection{Contact-implicit Approach with Differential Dynamic Programming}
The integration of the contact-implicit approach with DDP-based algorithms has been explored as in~\citet{Yuval2012, TO-w-contact, carius2018trajectory, chatzinikolaidis2021trajectory, kurtz2022contact, kong2023hybrid}. A key distinction from direct transcription methods is the treatment of the contact impulse, which becomes a function of both the state and control input. The complementarity constraint, causing numerical challenges in optimization, does not appear as an explicit constraint in the optimal control problem; instead, it is integrated into the system dynamics. Therefore, in the forward pass, the contact impulse, corresponding to the step's state and control input, needs to be computed, while in the backward pass, computation of the contact impulse's gradient is required. 

To facilitate the discovery of other contact modes, the gradient of the contact impulse is commonly computed using a differentiable contact model, as in~\citet{Yuval2012}. Introducing virtual forces on potential contact points aids in unveiling new contacts~\citep{onol2019contact,onol2020tuning}. Adopting a soft contact model with a smoothing term~\citep{todorov2011convex,todorov2014convex} further enhances this procedure, noticeable in MuJoCo MPC~\citep{howell2022predictive} and DDP with implicit dynamics formulation~\citep{chatzinikolaidis2021trajectory}. This approach can also be extended to MPC scenarios, demonstrated in simulations of humanoid getting-up and walking motions~\citep{Yuval2012}. 

\begin{figure*}
    \centering
    \includegraphics[width=2.0\columnwidth]{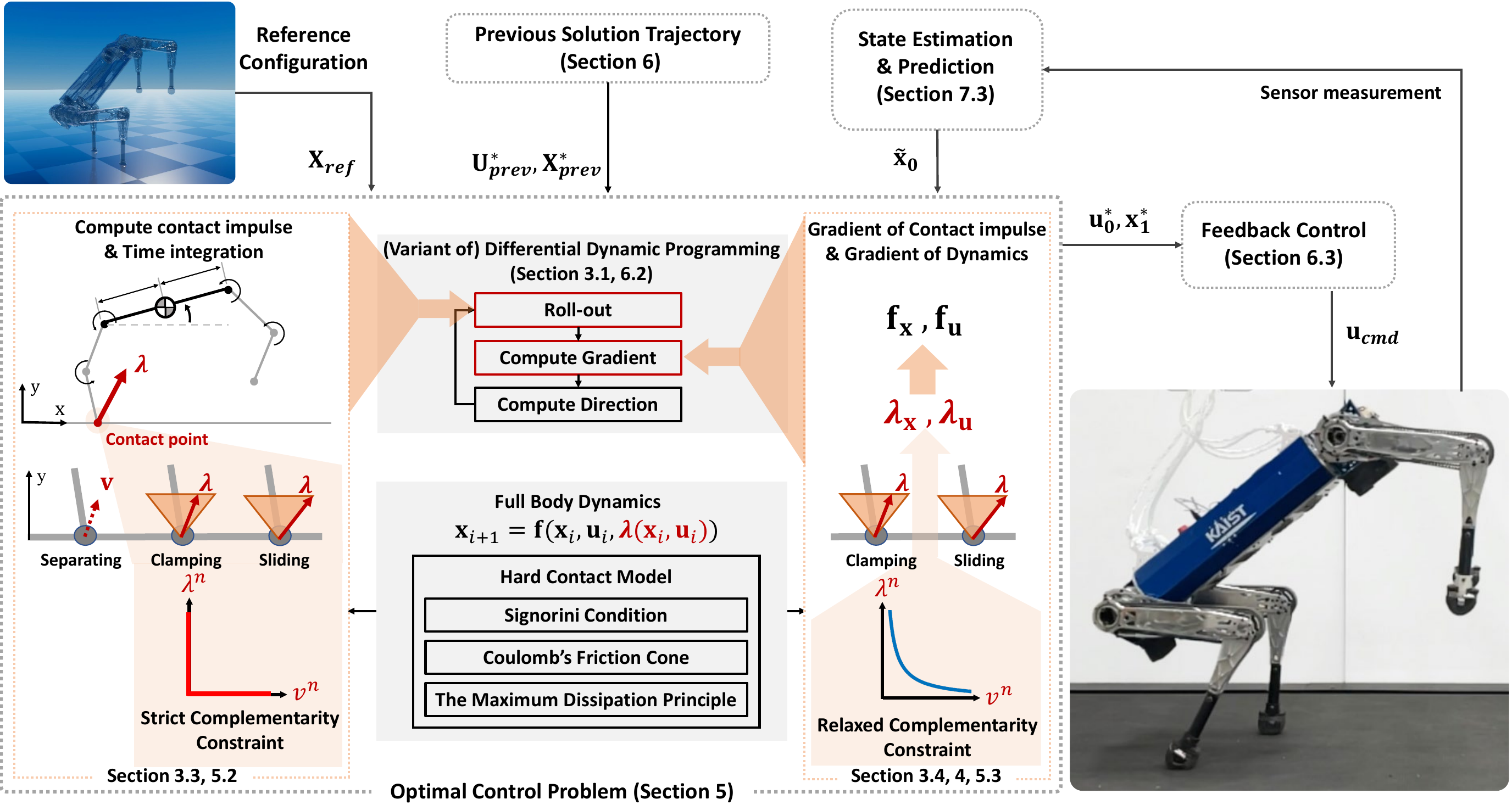}
    \caption[Whole framework of the contact-implicit DDP]{The overall framework of the contact-implicit model predictive control (MPC): The optimal control problem is tackled using a variant of the differential dynamic programming (DDP) algorithm, with the integration of a hard contact model within the system dynamics. The strict complementarity constraint, arising from the Signorini condition, is exclusively employed during the roll-out process, while the relaxed complementarity constraint is employed when computing the gradient component.}
    \label{figure:whole}
\end{figure*}

The contact-implicit DDP approach is emerging in quadruped robotics and real-hardware experiments, employing the spring-damper model for smooth gradients, as in ~\citet{TO-w-contact,neunert2018whole}. That compliant contact model is helpful to reason about the new contact modes while losing physical realism (e.g., ground penetration or non-vanishing contact forces). \citet{TO-w-contact} reveal diverse contact-rich motions, though versatile motion requires careful cost tuning and faces challenges in real-world transfer due to robustness issues. However, the MPC framework~\citep{neunert2018whole} enhances robustness, as demonstrated in hardware experiments with trotting and squat jump motions. Leveraging the contact-implicit approach, the method adapts contact sequences during trot motions to handle disturbances. While these previous approaches explore a diverse array of motions in real experiments, such as trotting and squat jumps, our implementation further extends the capabilities of real-world experiment to include challenging dynamic motions, such as front-leg rearing. 

Due to issues with the spring-damper model in \citep{neunert2018whole}, such as physically unrealistic motion and requiring small time steps due to numerical stiffness, a hard contact model was adopted \citep{carius2018trajectory, carius2019trajectory}. The contact impulse is computed through quadratically constrained quadratic programming (QCQP), while gradient computation relies on automatic differentiation. Optimized trajectories, including contact switching motions, are demonstrated through experiments involving one-leg hopper jumping~\citep{carius2018trajectory}, and quadruped robot walking and slippage motions~\citep{carius2019trajectory}. However, challenges persist, particularly concerning the discovery of new contact modes and the implementation of MPC under the cases of contact mismatches~\citep{carius2019trajectory}. 

\subsubsection{Proposed Method.}
Our approach, based on contact-implicit DDP with a hard contact model, reduces the smoothing effect on physical constraints (e.g., force at a distance) in simulation. The hard contact model follows the QCQP formulation with an explicit Signorini condition constraint, as in~\citet{raisim}, and the contact impulse gradient is analytically computed based on the approach in~\citet{KarenLiu}. To handle the challenges of exploring new contact modes, we introduce a novel smooth gradient based on the relaxed complementarity constraints: Exact simulation (no relaxation) in the forward pass and smooth gradient (with relaxation) in the backward pass, as depicted in Figure~\ref{figure:whole}. The problem of the unstable initial roll-out in the MPC setting is addressed by leveraging Feasibility-driven Differential Dynamic Programming (FDDP)~\citep{mastalli2020crocoddyl}, which is a multiple shooting variant of DDP. Additionally, we incorporate extra cost terms to enhance motion reliability, mitigating issues such as scuffing motion found in~\citet{carius2019trajectory}.

\section{Background}
\label{sec:background}
This section outlines the theoretical background of our framework. The optimal control problem is tackled using a DDP-based algorithm. To address the unstable initialization problem, FDDP, a multiple-shooting variant of DDP, is employed. Brief introductions to both DDP and its variant, FDDP, are included. Additionally, notation for full-body dynamics and contact dynamics is presented. Given the contact-implicit DDP framework, this section details the computation of the contact impulse and its analytical gradient, necessary for the DDP forward pass and backward pass, respectively.

\subsection{Differential Dynamic Programming}
\label{sec:DDP}
DDP~\citep{mayne1966second,Yuval2012} is a direct single-shooting approach to solve a finite-horizon optimal control problem with a discrete-time dynamical system: 
\begin{subequations}
\label{eq:DDP}
\begin{align}
    \min_{\mathbf{U}} ~ & l_N(\mathbf{x}_N) +\sum_{i=0}^{N-1} l(\mathbf{x}_i,\mathbf{u}_i) \\
    s.t. ~ & \mathbf{x}_{i+1}=\mathbf{f}(\mathbf{x}_i,\mathbf{u}_i), \label{eq:dynamics1}
\end{align} \end{subequations} 
where \(\mathbf{U} = \{\mathbf{u}_0, \ldots, \mathbf{u}_{N-1}\}\) represents sequence of control inputs, and \(l(\mathbf{x}_i, \mathbf{u}_i)\) and \(l_N(\mathbf{x}_N)\) are the running and final costs, respectively. \(\mathbf{x}_i\) and \(\mathbf{u}_i\) denote the state and control at time step \(i\), and \(\mathbf{f}\) is the generic function of dynamics. 

For this optimization problem~\eqref{eq:DDP}, the value function $V(\mathbf{x}_i)$ could be defined as optimal cost-to-go from $\mathbf{x}_i$. Based on Bellman's Optimality Principle, the following recursive relation holds: $V(\mathbf{x}_i) = \min_{\mathbf{u}_i}[l(\mathbf{x}_i,\mathbf{u}_i)+V'(\mathbf{f}(\mathbf{x}_i,\mathbf{u}_i))]$. Introducing the action-value function, which is defined as $Q(\mathbf{x}_i,\mathbf{u}_i)\triangleq l(\mathbf{x}_i,\mathbf{u}_i)+V'(\mathbf{f}(\mathbf{x}_i,\mathbf{u}_i))$, we approximate it locally in a quadratic manner to derive the control sequence improvement. The coefficients for this quadratic approximation are as follows:
\begin{equation}
\label{eq:coeff_of_Qfunction}
\begin{aligned}
    & Q_{\mathbf{x}} = l_{\mathbf{x}} + \mathbf{f}_{\mathbf{x}}^TV_{\mathbf{x}}',\quad Q_{\mathbf{u}}  = l_{\mathbf{u}} + \mathbf{f}_{\mathbf{u}}^TV_{\mathbf{x}}', \\
    & Q_{\mathbf{xx}} = l_{\mathbf{xx}} + \mathbf{f}_{\mathbf{x}}^TV_{\mathbf{xx}}'\mathbf{f}_{\mathbf{x}} + V_{\mathbf{x}}'\cdot \mathbf{f}_{\mathbf{xx}}, \\
    & Q_{\mathbf{xu}} = l_{\mathbf{xu}} + \mathbf{f}_{\mathbf{x}}^TV_{\mathbf{xx}}'\mathbf{f}_{\mathbf{u}} + V_{\mathbf{x}}'\cdot \mathbf{f}_{\mathbf{xu}}, \\
    & Q_{\mathbf{uu}} = l_{\mathbf{uu}} + \mathbf{f}_{\mathbf{u}}^TV_{\mathbf{xx}}'\mathbf{f}_{\mathbf{u}} + V_{\mathbf{x}}'\cdot \mathbf{f}_{\mathbf{uu}}. 
\end{aligned}     
\end{equation}
The action-value function $Q$, value function $V$, cost $l$ and dynamics function $\mathbf{f}$, with subscripts $\mathbf{x}$ and $\mathbf{u}$, represent the derivatives of each function with respect to the subscripted variables. Note that in this work, we do not compute the second derivative of dynamics $\mathbf{f}_{\mathbf{xx}}$, $\mathbf{f}_{\mathbf{xu}}$, and $\mathbf{f}_{\mathbf{uu}}$, corresponding to the Gauss-Newton approximation, such as used in the iterative Linear-Quadratic Regulator (iLQR)~\citep{li2004iterative}.

In the backward pass, minimizing the quadratic approximation of the action-value function yields the optimal control change as $\delta\mathbf{u}^*(\delta\mathbf{x}) = \mathbf{k}+\mathbf{K}\delta\mathbf{x}$, where the feed-forward term is \(\mathbf{k} \triangleq -Q_{\mathbf{uu}}^{-1}Q_{\mathbf{u}}\), and the feedback gain is \(\mathbf{K} \triangleq -Q_{\mathbf{uu}}^{-1}Q_{\mathbf{ux}}\). By utilizing it, we can compute the quadratic approximation of the value function, thereby establishing a recursive relation to determine the optimal control modification across all time steps, from \(N\) to 0.

In the forward pass, the states and control inputs are updated starting from the initial state \(\mathbf{\tilde{x}}_0\), using \(\mathbf{k}\) and \(\mathbf{K}\) from the backward pass:
\begin{equation}
\label{eq:forward}
\begin{aligned}
    \mathbf{\hat{x}}_0 &= \mathbf{\tilde{x}}_0, \\ 
    \mathbf{\hat{u}}_i &= \mathbf{u}_i+\alpha\mathbf{k}_i+\mathbf{K}_i(\mathbf{\hat{x}}_i \ominus \mathbf{x}_i),\\
    \mathbf{\hat{x}}_{i+1} &= \mathbf{f}(\mathbf{\hat{x}}_i,\mathbf{\hat{u}}_i), 
\end{aligned}    
\end{equation}
where $\ominus$ denotes the \textit{difference} operator of the state manifold (following the notations in~\citet{mastalli2022feasibility}), needed to optimize over the manifold~\citep{gabay1982minimizing}. $\alpha$ is the step size for line search, and regularization is necessary in practical applications. This method could be extended to address box constraints on control input~\citep{Yuval} and nonlinear state-control constraints~\citep{KRIS_CDDP}.

\subsubsection{Feasibility-driven Differential Dynamic Programming.}
FDDP~\citep{mastalli2020crocoddyl} is a multiple-shooting variant of DDP, offering enhanced globalization compared to classical DDP. It accommodates infeasible state-control trajectories for warm-start initialization by adjusting the backward pass and opening the dynamics gaps in early iterations in the forward pass.

For the infeasible state-control trajectory $\mathbf{x}_0,...,\mathbf{x}_{N}$, $\mathbf{u}_0,...,\mathbf{u}_{N-1}$, where $\mathbf{x}_i$ represents the shooting state, the dynamics gap $\mathbf{\bar{f}}_{i}$ is determined as follows: 
\begin{equation}
\label{eq:fs_gap}
\begin{aligned}
    & \mathbf{\bar{f}}_0 := \mathbf{\tilde{x}}_0 \ominus \mathbf{x}_0, \\ 
    & \mathbf{\bar{f}}_{i+1} := \mathbf{f}(\mathbf{x}_i,\mathbf{u}_i) \ominus \mathbf{x}_{i+1}, \quad{\scriptstyle \forall i=\{0,1,...,N-1\}},
\end{aligned}    
\end{equation}
where $\mathbf{\tilde{x}}_0$ denotes the initial state, and $\mathbf{\bar{f}}_{0},\mathbf{\bar{f}}_{i+1}$ indicate the dynamics gap, on the tangent space of the state manifold. 

The backward pass is adapted to accept infeasible trajectories as proposed in~\citet{giftthaler2018family}. It adjusts the Jacobian of the value function to reflect the deflection from the gap as $V_{\mathbf{x}}^+ = V_{\mathbf{x}}' + V_{\mathbf{xx}}' \mathbf{\bar{f}}_{i+1}$. Subsequently, \( V_{\mathbf{x}}^+ \) is utilized to compute~\eqref{eq:coeff_of_Qfunction} instead of $V_{\mathbf{x}}'$.

In the forward pass, the nonlinear roll-out process for $\alpha$-step is given by:
\begin{equation}
\label{eq:forward_FDDP}
\begin{aligned}
    \mathbf{\hat{x}}_0 &= \mathbf{\tilde{x}}_0 \oplus (\alpha-1)\mathbf{\bar{f}}_0, \\ 
    \mathbf{\hat{u}}_i &= \mathbf{u}_i+\alpha\mathbf{k}_i+\mathbf{K}_i(\mathbf{\hat{x}}_i \ominus \mathbf{x}_i),\\
    \mathbf{\hat{x}}_{i+1} &= \mathbf{f}(\mathbf{\hat{x}}_i,\mathbf{\hat{u}}_i) \oplus (\alpha-1)\mathbf{\bar{f}}_{i+1}, 
\end{aligned}    
\end{equation}
where $\oplus$ denotes the \textit{integrator} operator, needed to optimize over the manifold~\citep{gabay1982minimizing}. This roll-out maintains an equivalent gap contraction rate for the \( \alpha \)-step in direct multiple shooting, given by \( \mathbf{\bar{f}}_i \leftarrow (1-\alpha) \mathbf{\bar{f}}_{i} \). FDDP approves the trial step in an ascent direction to decrease infeasibility while slightly increasing the cost. Due to its feasibility-driven approach, FDDP exhibits a larger basin of attraction to the good local optimum than DDP~\citep{mastalli2022feasibility}. This method has been extended to address control limits as presented in Box-FDDP~\citep{mastalli2022feasibility}, which serves as the primary optimal control problem solver in our work.

\subsection{Notation for Full Body Dynamics and Contact Dynamics}
In this section, we focus our interest on the case when the discrete-time dynamic model~\eqref{eq:dynamics1} incorporates contact dynamics. Our framework employs a velocity-based time-stepping scheme.
 Given contact impulse $\bm{\lambda}_i$ at $i$-th time step, next time step's generalized coordinate $\mathbf{q}_{i} \in \mathrm{SE}(3)\times\mathbb{R}^{n_j}$ and generalized velocity $\mathbf{\dot{q}}_{i} \in \mathfrak{se}(3)\times\mathbb{R}^{n_j}$ can be obtained:
\begin{equation}
\label{eq:dynamics}
\begin{aligned}
    & \mathbf{q}_{i+1} =\mathbf{q}_{i}\oplus\mathbf{\dot{q}}_{i+1}dt, \\ 
    & \mathbf{\dot{q}}_{i+1} = \mathbf{M}_{i}^{-1}((-\mathbf{h}_{i}+\mathbf{B}\mathbf{u}_i)dt + \mathbf{M}_{i}\mathbf{\dot{q}}_i+\mathbf{J}_{i}^T\bm{\lambda}_{i}),
\end{aligned} 
\end{equation}
where $\mathbf{\dot{q}}_{i}$ is the tangent vector to the configuration $\mathbf{q}_{i}$ with dimension $n_{\dot{q}}$ (not the same dimension with $\mathbf{q}_{i}$). $\mathbf{u}_i \in \mathbb{R}^{n_j}$ are the joint torque, and $n_j$ is the number of articulated joints. $\mathbf{M}_i =\mathbf{M}(\mathbf{q}_{i})\in\mathbb{R}^{n_{\dot{q}}\times n_{\dot{q}}}$ is the mass matrix, $\mathbf{h}_i=\mathbf{h}(\mathbf{q}_i,\mathbf{\dot{q}_i})\in\mathbb{R}^{n_{\dot{q}}}$ is a bias vector including Coriolis and gravitational terms, $\mathbf{B}=[\mathbf{0}_{n_j\times 6},\mathbf{I}_{n_j \times n_j}]^T \in \mathbb{R}^{n_{\dot{q}} \times n_{j}}$ is an input matrix, and $dt$ is a time step. 

Here, $\bm{\lambda}_{i}=[\bm{\lambda}_{1,i}^T,\cdots,\bm{\lambda}_{d,i}^T]^T$ denotes the concatenation of all contact impulses, and $\bm{\lambda}_{k,i} \in \mathbb{R}^3$ denotes the contact impulse of the $k$-th contact point (among $d$ total contact points) in the contact frame, with normal and tangential components ${\lambda}^n_{k,i}$ and ${\bm{\lambda}}^t_{k,i}$, respectively. The superscripts $n$ and $t$ represent normal and tangential components.

Similarly, the contact Jacobian $\mathbf{J}_{i}=[\mathbf{J}_{1,i}^T, \cdots \mathbf{J}_{d,i}^T]^T$ comprises of individual contact Jacobians for each point, with $\mathbf{J}_{k,i} \in \mathbb{R}^{3\times n_{\dot{q}}}$. The resulting contact velocity for the $k$-th point at time step $i+1$, $\mathbf{v}_{k,i+1}=\mathbf{J}_{k,i}\mathbf{\dot{q}}_{i+1}$, encompasses the contact impulse's effect from time $i$ and  is split into normal $v^n_{k,i+1}$ and tangential $\mathbf{v}^t_{k,i+1}$ components. 

For brevity in Sections~\ref{sec:LCP}, \ref{sec:contact_dyn}, \ref{sec:gradients}, and \ref{sec:analytical_gradient_based_on_relaxation}, the subscript time index \( i \) is omitted for variables such as \( \mathbf{M}, \mathbf{J}_{k}, \mathbf{h} \), and \( \bm{\lambda}_k \). However, for clarity, \( i \) is retained for \( \mathbf{\dot{q}}_{i} \) and \( \mathbf{u}_{i} \). Additionally, the time index for the next time step, \( i+1 \), is consistently denoted (e.g., \( \mathbf{v}_{k,i+1} \), \( \mathbf{\dot{q}}_{i+1} \)). Table~\ref{tab:notation} contains notations for key variables.

\begin{table}[htbp]
\centering
\caption{Notation of variables (Nomenclature).}
\label{tab:notation}
\begin{tabular}{|c|c|}
\hline
\textbf{Name} & \textbf{Description}\\
\hline
$(\cdot)_i$ & index for $i$-th time step\\
$(\cdot)_k$ & index for $k$-th contact point (foot index)\\
\hline
$(\cdot)^n$ & normal component\\
$(\cdot)^t$ & tangential component\\
\hline
$\mathbf{q}$ & generalized coordinate\\
$\dot{\mathbf{q}}$ & generalized velocity\\
$\mathbf{u}$ & control input (joint torque)\\
\hline
$\mathbf{J}$ & contact Jacobian\\
$\mathbf{v}$ & contact velocity\\
$\bm{\lambda}$ & contact impulse\\
$\phi$ & signed distance to contact point (foot height)\\
\hline
$\rho$ & relaxation variable\\
\hline
$\mathbf{x}$ & state, $\mathbf{x}=(\mathbf{q}, \mathbf{\dot{q}})$ \\
$\mathbf{\hat{x}}$ & roll-out state\\
$\mathbf{\tilde{x}}_0$ & initial state (estimated state) \\
$\mathbf{\bar{f}}$ & dynamics gap \\ 
\hline
\end{tabular}
\end{table}

\subsubsection{Linear Complementarity Constraint.}
\label{sec:LCP}
The Signorini condition for hard contact is expressed as $\phi_{k,i+1}\geq 0,\lambda_{k}^n\geq 0,\text{and } \phi_{k,i+1}\lambda_{k}^n=0$, where $\phi_{k,i+1}$ is the signed distance from the $k$-th contact point to the surface. In this condition, the non-contacting point maintaining a distance from the surface does not receive a contact impulse, while only the contacting point does. For computational simplicity, the Signorini condition can be expressed in velocity space as $v^{n}_{k,i+1}\geq 0,\lambda_{k}^n\geq 0,\text{and } v^{n}_{k,i+1}\lambda_{k}^n=0$. 

Given the time-stepping scheme (equation~\eqref{eq:dynamics}), the contact velocity $\mathbf{v}_{k,i+1}$ is an affine function of the contact impulse $\bm{\lambda}_{k}$. This formulation allows the Signorini condition in velocity space to be represented as a linear complementarity constraint. In the following subsection, this constraint is utilized to compute the feasible contact impulse.

\subsection{Contact Impulse}
\label{sec:contact_dyn}
For the integration step of roll-out~\eqref{eq:forward}, contact impulse $\bm{\lambda}$ is obtained via per-contact iteration method with the hard contact model in~\citet{preclik2014models,raisim}. The contact scenarios can be categorized based on Signorini condition in velocity space. For opening contact (separating), the contact impulse is zero. For closing contact (clamping and sliding), the contact impulse for the $k$-th contact point is computed by solving the following optimization problem that aims to minimize the kinetic energy at the contact point, based on the maximum dissipation principle:
\begin{equation}
\label{eq:minvel}
\begin{aligned}
    \min_{\bm{\lambda}_k}{\mathbf{v}_{k,i+1}}^T \mathbf{M}^{\text{app}}_{k}{\mathbf{v}_{k,i+1}}\\
    s.t. \quad \bm{\lambda}_{k}\in {\cal{S}}_k,    
\end{aligned}
\end{equation}
where $\mathbf{M}^{\text{app}}_{k}:=(\mathbf{J}_k \mathbf{M}^{-1} \mathbf{J}_k^T)^{-1}$ is the apparent inertia matrix at the $k$-th contact point. Given that $\tilde{\mathbf{k}}$ denotes all the contact indices only except $k$, the contact velocity of the $k$-th contact point can be obtained by:
\begin{align}
\label{eq:contactvel}
    \mathbf{v}_{k,i+1}=\mathbf{c}_k+(\mathbf{M}^{\text{app}}_{k})^{-1} \bm{\lambda}_k,
\end{align}
where $\mathbf{c}_k:=\mathbf{J}_k \mathbf{M}^{-1} ((-\mathbf{h}+\mathbf{B} \mathbf{u}_i)dt +\mathbf{J}_{\tilde{\mathbf{k}}}^T \bm{\lambda}_{\tilde{\mathbf{k}}}+\mathbf{M} \mathbf{\dot{q}}_i)$. 
The feasible set ${\cal{S}}_k$ is formed by the following conditions:
\begin{subequations}
\begin{align}
v^{n}_{k,i+1}\geq 0,~~\lambda_{k}^n&\geq 0,~~\text{and } v^{n}_{k,i+1}\lambda_{k}^n=0 \label{eq:condition1},  \\
{\| \bm{\lambda}}^t_k\|&\leq \mu \lambda^n_k \label{eq:condition2},
\end{align}
\end{subequations}
where $\mu$ is the friction coefficient. The condition~\eqref{eq:condition1} arises from the Signorini condition, while the constraint~\eqref{eq:condition2} represents Coulomb's friction cone constraint. Currently addressing closing contact scenarios (clamping and sliding), the condition~\eqref{eq:condition1} simplifies to an affine condition as $v^{n}_{k,i+1}=0$. In the clamping case, the optimal solution is $\bm{\lambda}_k=-\mathbf{M}_{k}\mathbf{c}_k$. For the sliding case, the optimization problem~\eqref{eq:minvel} is formulated to QCQP~\citep{lidec2023contact}, solvable via the bisection method~\citep{raisim}.   

In scenarios with multiple contacts, the contact impulses $\bm{\lambda}_1, \cdots, \bm{\lambda}_d$ for all contact points can be obtained by solving the optimization problem~\eqref{eq:minvel} for all $k=1,\cdots,d$, iteratively. 

\subsection{Analytical Gradient of Contact Impulse}
\label{sec:gradients}
\begin{figure}
    \centering
    \includegraphics[width=0.8\columnwidth]{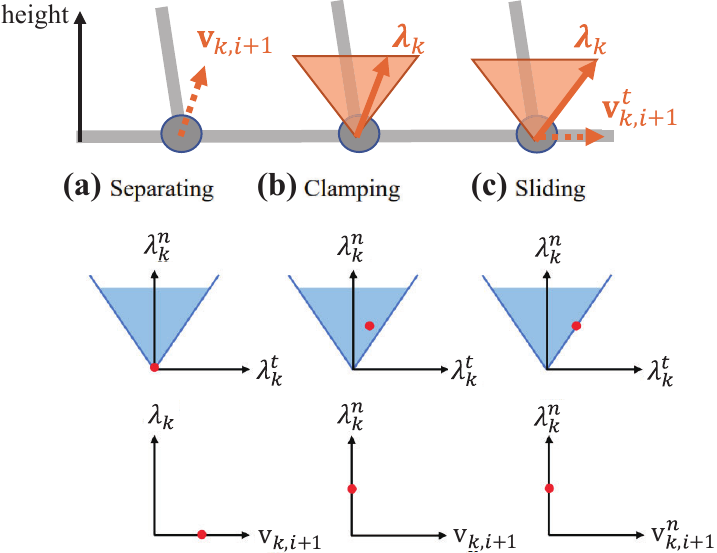}
    \caption[Graphical illustration about three contact cases]{Illustration of a contact impulse and contact velocity in the cases of (a) Separating, (b) Clamping, and (c) Sliding}
    \label{figure:contactforces}
\end{figure}

To compute the derivatives of the dynamics $\mathbf{f}_{\mathbf{x}}$ and $\mathbf{f}_{\mathbf{u}}$ for the backward pass detailed in~\eqref{eq:coeff_of_Qfunction}, the gradient of the contact impulse is required. Based on~\citet{KarenLiu}, the analytical gradient of the contact impulse is computed. The solution of contact impulse can be categorized into three types according to the value of $\bm{\lambda}_k$ and $\mathbf{v}_{k,i+1}$ as shown in Figure~\ref{figure:contactforces}, due to the Signorini condition and Coulomb's friction constraint. Along this categorization, we apply specific conditions for gradient computation that are unique to each contact type.

\subsubsection{Separating.} The gradient of $\frac{\partial {\bm{\lambda}}_k}{\partial \xi}$ is zero for an arbitrary variable $\xi$ because $\bm{\lambda}_k=\bm{0}$.

\subsubsection{Clamping.} With ${v}^{n}_{k,i+1} =0$ and $\mathbf{v}^{t}_{k,i+1}={0}$, the contact point remains contact with the ground, not sliding in the tangential direction at the next time step. The gradient of $\frac{\partial \bm{\lambda}_k}{\partial \xi}$ has to be obtained using the constraints $\mathbf{v}_{k,i+1}=0$.

\subsubsection{Sliding.} Given ${v}^{n}_{k,i+1} =0$, the contact point remains contact with the ground, sliding with tangential velocity $\mathbf{v}^{t}_{k,i+1}$. The contact impulse exists on the boundary of the friction cone, expressed as $\bm{\lambda}_k=\mathbf{E}_k{\lambda^n_k}$, where the matrix $\mathbf{E}_k = [\mu cos(\theta_k), \mu sin(\theta_k), 1]^T$ is determined through the optimization problem in~\eqref{eq:minvel}. The gradient of $\frac{\partial \bm{\lambda}_k}{\partial \xi}$ has to be obtained using the constraints ${v}^{n}_{k,i+1} =0$ and $\bm{\lambda}_k=\mathbf{E}_k{\lambda^n_k}$.

To compute $\frac{\partial \bm{\lambda}_k}{\partial \xi}$  for clamping and sliding scenarios, let $\mathbf{c}$ denote the set of indices for clamping contact points and $\mathbf{s}$ for sliding contact points. Any variable $\mathbf{y}$ with the subscript $\mathbf{c}$ or $\mathbf{s}$ will denote a vector or matrix constructed by vertically concatenating  $\mathbf{y}_c$ or $\mathbf{y}_s$ for all $c\in\mathbf{c}$ or $s\in\mathbf{s}$. For example, given $\mathbf{c}=\{1,3\}$, $\mathbf{J}_{\mathbf{c}}=\left[\mathbf{J}_1^T,~\mathbf{J}_3^T\right]^T$ and $\mathbf{v}_{\mathbf{c},i+1}=\left[{\mathbf{v}_{1,i+1}}^T,~{\mathbf{v}_{3,i+1}}^T\right]^T$. In the sliding contact cases, the matrix $\mathbf{E}_{\mathbf{s}}$ is a block diagonal matrix with top-left entry $\mathbf{E}_{s_1}$ and bottom-right entry $\mathbf{E}_{s_n}$, where $s_1$ and $s_n$ are the first and last elements of set $\mathbf{s}$, respectively.

Then, for clamping (set $\mathbf{c}$) and sliding (set $\mathbf{s}$) contact points, the following algebraic equations should hold:
\begin{equation}
\label{eq:equality}
\begin{aligned}
    \mathbf{v}_{\mathbf{c},i+1}=\mathbf{0}, \\    
    \mathbf{v}^{n}_{\mathbf{s},i+1}=\mathbf{0},
\end{aligned}        
\end{equation}
where $\mathbf{v}_{\mathbf{c},i+1}$ is a vector from the space $\mathbb{R}^{3n_{\mathbf{c}}}$, and $\mathbf{v}^{n}_{\mathbf{s},i+1}$ is from the space $\mathbb{R}^{n_{\mathbf{s}}}$, with $n_{\mathbf{c}}$ and $n_{\mathbf{s}}$ representing the number of clamping points and sliding points, respectively.

From the equation~\eqref{eq:dynamics}, $\mathbf{\dot{q}}_{i+1}$ can be obtained as follows, considering $\bm{\lambda}_{\mathbf{s}}=\mathbf{E}_{\mathbf{s}}{\bm{\lambda}^n_{\mathbf{s}}}$ for sliding cases:
\begin{align}
    \mathbf{\dot{q}}_{i+1}\nonumber=\mathbf{M}^{-1}((-\mathbf{h}+\mathbf{B} \mathbf{u}_i)dt+\mathbf{M}\mathbf{\dot{q}}_i+{\mathbf{J}_{\mathbf{c}}}^T \bm{\lambda}_{\mathbf{c}}+{\mathbf{J}_{\mathbf{s}}}^T \mathbf{E}_{\mathbf{s}}\bm{\lambda}^n_{\mathbf{s}})\nonumber.
\end{align}

Identifying $\mathbf{v}_{\mathbf{c},i+1}=\mathbf{J}_{\mathbf{c}}\mathbf{\dot{q}}_{i+1}$ and $\mathbf{v}^{n}_{\mathbf{s},i+1}=\mathbf{J}^{n}_{\mathbf{s}}\mathbf{\dot{q}}_{i+1}$, and then vertically concatenating two equations of \eqref{eq:equality}, we can rewrite \eqref{eq:equality} as,
\begin{align}
\label{eq:Acc}
    \mathbf{0}=\mathbf{A}\bm{\lambda}^{\text{contact}}+\mathbf{b},
\end{align}
where, 
\begin{align}
\bm{\lambda}&^{\text{contact}}=\begin{bmatrix}
{\bm{\lambda}_{\mathbf{c}}}^T & {\bm{\lambda}^n_{\mathbf{s}}}^T
\end{bmatrix}^T,\nonumber\\
    \mathbf{A}&=\begin{bmatrix}
    \mathbf{J}_{\mathbf{c}}\nonumber\\\mathbf{J}^n_{\mathbf{s}}
    \end{bmatrix}
    \mathbf{M}^{-1}
    \begin{bmatrix}
    \mathbf{J}_{\mathbf{c}}\\\mathbf{E}_{\mathbf{s}}^T{\mathbf{J}_{\mathbf{s}}}
    \end{bmatrix}^T,\nonumber\\
    \mathbf{b}&=\begin{bmatrix}
    \mathbf{J}_{\mathbf{c}}\\\mathbf{J}^n_{\mathbf{s}}
    \end{bmatrix}\mathbf{M}^{-1}\left((-\mathbf{h}+\mathbf{B} \mathbf{u}_{i})dt + \mathbf{M}\mathbf{\dot{q}}_i\right)\nonumber,
\end{align}
where $\mathbf{A}$ represents the Delassus matrix, and $\mathbf{b}$ denotes the contact velocity of the unconstrained motion, excluding the influence of contact impulses. 

With~\eqref{eq:Acc}, the contact impulse $\bm{\lambda}^{\text{contact}}$ is represented as,
\begin{align}
\label{eq:lambda}
    \bm{\lambda}^{\text{contact}}=-\mathbf{A}^{-1}\mathbf{b}.
\end{align}
Then, the gradient of $\bm{\lambda}^{\text{contact}}$ with respect to $\xi$ is obtained:
\begin{align}
\label{eq:nonrelaxed_gradient}
    \frac{\partial \bm{\lambda}^{\text{contact}}}{\partial \xi}= \mathbf{A}^{-1} \frac{\partial \mathbf{A}}{\partial \xi}\mathbf{A}^{-1}\mathbf{b}-\mathbf{A}^{-1}\frac{\partial \mathbf{b}}{\partial \xi}.
\end{align}
Furthermore, utilizing $\bm{\lambda}_{\mathbf{s}}=\mathbf{E}_{\mathbf{s}}{\bm{\lambda}^n_{\mathbf{s}}}$, the gradient of $\bm{\lambda}_{\mathbf{s}}$ including tangential direction can be obtained by,
\begin{align}
\label{eq:nonrelaxed_gradient2}
    \frac{\partial\bm{\lambda}_{\mathbf{s}}}{\partial \xi}=\mathbf{E}_{\mathbf{s}}\frac{\partial\bm{\lambda}^n_{\mathbf{s}}}{\partial \xi}.
\end{align}
Equations~\eqref{eq:nonrelaxed_gradient} and~\eqref{eq:nonrelaxed_gradient2} are used to obtain $\frac{\partial \bm{\lambda}}{\partial \mathbf{q}_i}$, $\frac{\partial \bm{\lambda}}{\partial \mathbf{\dot{q}}_i}$, and $\frac{\partial \bm{\lambda}}{\partial \mathbf{u}_i}$. 

\section{Analytical Gradient based on Relaxed Complementarity Constraints}
\label{sec:analytical_gradient_based_on_relaxation}

\subsection{Problem Statement}
The analytical gradient in Section~\ref{sec:gradients} is computed based on established contact modes during the roll-out, governed by the hard contact model. As the gradient is computed in a local region that preserves the contact state, it offers limited information to guide other contact modes as described in~\citet{KarenLiu}.

For clamping cases, the gradient is computed for the local region where $v^{n}_{k,i+1}=0$, encouraging maintenance of a zero contact velocity. As a result, this search direction inhibits motions for breaking contact to lower costs. For example, when initialized in a fully-contacted state, such as a quadruped's standing motion, the optimizer tends to maintain and converge to the fully-contacted state. Consequently, cost reduction is restricted when further decreases necessitate breaking contact for actions such as jumping to reach a distant reference configuration.

The same holds true for separating cases. However, while the gradient encourages a zero contact impulse, gravitational effects present opportunities for making new contact instances. For example, a falling robot without any active control would eventually establish contact, resulting in the emergence of new contacts during the roll-out process. This is also pointed out in~\citet{tassa2010stochastic}. However, for the clamping scenarios, there is no inherent force that guides the system towards breaking the contact.

\begin{figure}
    \centering
    \includegraphics[width=0.66\columnwidth]{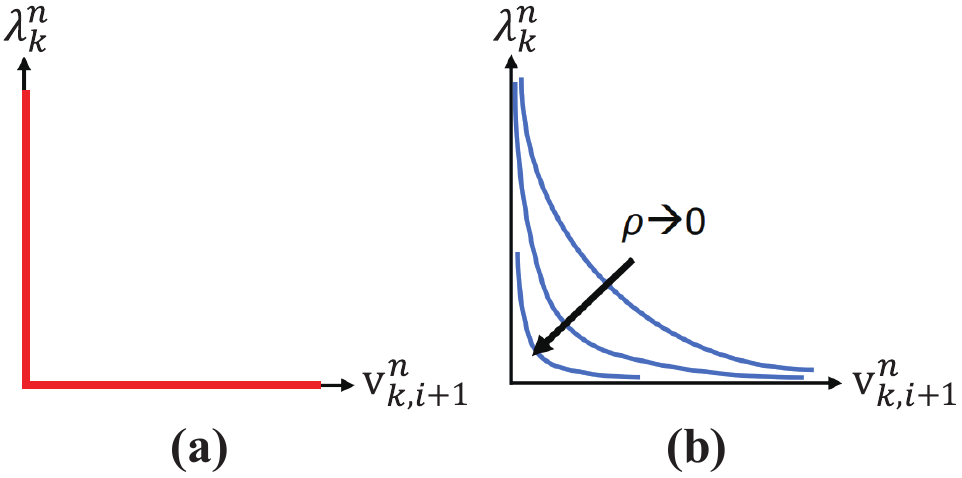}
    \caption[Graphical illustration about strict linear complementarity condition and relaxed complementarity condition]{Illustration of (a) a Complementarity condition and (b) relaxed Complementarity condition with the parameter $\rho$}
    \label{figure:contactforces2}
\end{figure}

The lack of information to the direction of breaking contact can be alleviated by relaxing the bilinear condition ${v}^{n}_{k,i+1}\lambda^{n}_{k} =0$ to ${v}^{n}_{k,i+1}\lambda^{n}_{k}=\rho$ for some small $\rho\geq 0$ as shown in Figure~\ref{figure:contactforces2}. This allows the solutions to move along the smooth curve ${v}^{n}_{k,i+1}{\lambda}^{n}_{k}=\rho$ during iterations rather than to stick to either ${v}^{n}_{k,i+1}=0$ or ${\lambda}^{n}_{k}={0}$. Such relaxation strategies have found successful applications across various optimization algorithms for trajectory optimization~\citep{Manchester2,patel} and quadratic programming~\citep{qpSWIFT}.

To integrate this relaxation into the contact-implicit DDP framework, we propose the analytical gradient of contact impulse based on the relaxed complementarity constraint. The other studies~\citep{Manchester2,patel,mordatch2012discovery} that \textit{relax} the physical constraint require the relaxation to reduce to zero through iterations for feasibility. However, our method employs this relaxed gradient only in the backward pass, and the forward roll-out is governed by a strict complementarity constraint. Therefore, in our approach, the use of relaxation does not compromise the trajectory's feasibility. By using the relaxed gradient, the optimizer can explore motions with contact breakage, transitioning from the initial contacted state to achieve refined solutions. 

\subsection{Gradient based on Relaxed Condition}
From the equation \eqref{eq:Acc}, the linear relationship between contact velocity and contact impulse is expressed as:
\begin{align}
    &\mathbf{v}^{\text{contact}}_{i+1} = \mathbf{A} \bm{\lambda}^{\text{contact}} + \mathbf{b} \label{velocity_relation},
\end{align}
where, $\mathbf{v}^{\text{contact}}_{i+1}=\left[ {\mathbf{v}_{\mathbf{c},i+1}}^T,~{\mathbf{v}^n_{\mathbf{s},i+1}}^T\right]^T $ of \eqref{eq:equality} and $\bm{\lambda}^{\text{contact}} = \begin{bmatrix}
{\bm{\lambda}_{\mathbf{c}}}^T,~ {\bm{\lambda}^n_{\mathbf{s}}}^T \end{bmatrix}^T $ of \eqref{eq:Acc}. 
For the normal velocities $\mathbf{v}^n_{\mathbf{c},i+1}$ and $\mathbf{v}^n_{\mathbf{s},i+1}$, complemetarity conditions are written as,
\begin{align}
    & {v}[l]\geq0 , ~~{\lambda}[l]\geq0,~~\text{and}~{v}[l]\lambda[l] = 0, \quad {\scriptstyle \forall l \in \mathcal{N}}, \nonumber
\end{align}
where ${v}[l]$ and $\lambda[l]$ are the $l$~th component of vectors $\mathbf{v}^{\text{contact}}_{i+1}$ and $\bm{\lambda}^{\text{contact}}$, respectively, where \( l \in \mathcal{N} \) with \( \mathcal{N} \) representing the set of indices corresponding to the normal direction components.
The constraints can be relaxed using a relaxation variable $\rho > 0$:
\begin{align}
{v}[l]>0 ,~~{\lambda}[l]>0,~~\text{and}~{v}[l]\lambda[l] = \rho, \quad{\scriptstyle \forall l \in \mathcal{N}}. \nonumber
\end{align}
Then each $l$ th row of \eqref{velocity_relation} can be rewritten as follows,
\begin{align}
    \frac{\rho}{\lambda[l]} = \mathbf{a}_l^T\bm{\lambda}^{\text{contact}} + b_l, \quad{\scriptstyle \forall l \in \mathcal{N}},
\end{align}    
where $\mathbf{a}_l^T$ is the $l$ th row of matrix $\mathbf{A}$, and $b_l$ is the $l$ th component of vector $\mathbf{b}$.

Differentiating this with respect to an arbitrary variable $\xi$ gives the implicit relation of $\frac{\partial \bm{\lambda}^{\text{contact}}}{\partial \xi}$ and $\bm{\lambda}^{\text{contact}}$:
\begin{align}
    -\frac{\rho}{\lambda[l]^2}&
    (\mathbf{e}_l^T\frac{\partial\bm{\lambda}^{\text{contact}}}{\partial\xi})\nonumber\\
    =& \frac{\partial \mathbf{a}_l^T}{\partial\xi}\bm{\lambda}^{\text{contact}}
    + \mathbf{a}_l^T\frac{\partial\bm{\lambda}^{\text{contact}}}{\partial\xi}
    +\frac{\partial b_l}{\partial\xi},
    \quad{\scriptstyle \forall l \in \mathcal{N}},
\end{align}
where $\mathbf{e}_l$ is the $l$ th elementary vector. Then, a system of linear equations is obtained:
\begin{align}
    \renewcommand\arraystretch{3}
    & \begin{bmatrix}  ...\\ \mathbf{a}_l^T+\frac{\rho}{\lambda[l]
    ^2}\mathbf{e}_l^T\\... \end{bmatrix} \frac{\partial\bm{\lambda}^{\text{contact}}}{\partial \xi} +\begin{bmatrix}  ...\\ \frac{\partial \mathbf{a}_l^T}{\partial \xi} \bm{\lambda}^{\text{contact}}+\frac{\partial b_l}{\partial \xi}\\... \end{bmatrix}=\mathbf{0}.
    \label{eq:relaxation_matrix_form}
\end{align}
In the case of tangential velocities, which are not subject to relaxation, the relation in \eqref{eq:relaxation_matrix_form} still holds with $\rho = 0$, 
by \eqref{eq:lambda} and \eqref{eq:nonrelaxed_gradient}.

Finally, we derive a simplified gradient equation:
\begin{align}
    &\frac{\partial \bm{\lambda}^{\text{contact}}}{\partial \xi}=-[\mathbf{A}+\rho \mathbf{D}]^{-1}(\frac{\partial \mathbf{A}}{\partial \xi} \bm{\lambda}^{\text{contact}}+\frac{\partial \mathbf{b}}{\partial \xi}), \label{eq:relaxation_matrix_form2}
\end{align}
where $\mathbf{D}$ is the diagonal matrix which has $l$~th diagonal element as $\frac{1}{\lambda[l]^2}$ if $l\in \mathcal{N}$, and 0 otherwise. It is noteworthy that when \(\rho = 0\), the derived gradient \(\frac{\partial \bm{\lambda}^{\text{contact}}}{\partial \xi}\) in~\eqref{eq:relaxation_matrix_form2} is identical to the gradient computed based on the strict constraint as expressed in \eqref{eq:nonrelaxed_gradient}.

This relaxed version of the gradient $\frac{\partial \bm{\lambda}^{\text{contact}}}{\partial \xi}$ is applied instead of \eqref{eq:nonrelaxed_gradient} to obtain the gradients of the contact impulse $\frac{\partial \bm{\lambda}}{\partial \mathbf{q}_i}$, $\frac{\partial \bm{\lambda}}{\partial \mathbf{\dot{q}}_i}$, and $\frac{\partial \bm{\lambda}}{\partial \mathbf{u}_i}$. 

\section{Contact-implicit Model Predictive Control}
\label{sec:motion_generation}
In the contact-implicit DDP framework, we integrate full-body robot dynamics with contact dynamics, solving the optimal control problem for the 3D case through Box-FDDP in Crocoddyl's solver~\citep{mastalli2020crocoddyl}, and for the 2D case through constrained DDP implemented in the DDP codebase \citep{tassa2015ilqg}. For the computation of the dynamics and its analytical derivatives, we use Pinocchio library~\citep{Pinocchio,carpentier2018analytical}. 

This framework is applied for a quadruped robot that has a single contact point per foot, \(d=4\). We model each foot contact as a point contact, rather than a surface contact. In all experiments and 3D simulations using the HOUND quadruped robot, the optimal control problem is discretized with a time step of \(dt = 25 \, \text{ms}\), has a horizon of \(N = 20\), and runs in an MPC fashion at 40 Hz.

\subsection{Cost}
In this paper, four types of costs are utilized: regulating cost $l_r(\mathbf{x}_i,\mathbf{u}_i)$, foot slip and clearance cost $l_f(\mathbf{x}_i,\mathbf{u}_i)$, air time cost $l_a(\mathbf{x}_i)$, and symmetric control cost $l_s(\mathbf{u}_i)$. The regulating cost alone enables multi-contact motion; however, we introduce three additional costs to ensure tractable motion. For the $i$-th time step, the running cost $l(\mathbf{x}_i,\mathbf{u}_i)$ and the final cost $l_N(\mathbf{x}_N)$ are specified as follows: 
\begin{align}
\label{eq:cost_total}
    & l(\mathbf{x}_i,\mathbf{u}_i) = l_r(\mathbf{x}_i,\mathbf{u}_i)+l_f(\mathbf{x}_i)+l_a(\mathbf{x}_i)+l_s(\mathbf{u}_i), \\
    & l_N(\mathbf{x}_N) = l_{r,N}(\mathbf{x}_N) \nonumber,
\end{align}
where $\mathbf{x}_{i}=(\mathbf{q}_{i},\mathbf{\dot{q}}_{i})$ is the robot state, $\mathbf{u}_i\in \mathbb{R}^{n_j}$ is the control input (joint torque), and $l_{r,N}(\mathbf{x}_N)$ is the regulating cost for terminal state. $l_r(\mathbf{x}_i,\mathbf{u}_i)$ and $l_f(\mathbf{x}_i,\mathbf{u}_i)$ are applied in all 3D simulations and experiments, while $l_a(\mathbf{x}_i)$ and $l_s(\mathbf{u}_i)$ are specifically used for encouraging symmetric gaits.

\subsubsection{Regulating Cost.} 
The regulating cost is a quadratic form designed to minimize the error between the desired and actual values for both configuration and control input:
\begin{align}
\label{eq:cost_regulating}
    &l_r(\mathbf{x}_i,\mathbf{u}_i) = \|\epsilon_{\mathbf{x}_i}\|_{\mathbf{W}_{\mathbf{x}}}^2 + \|\mathbf{u}_i-\mathbf{u}_{i,\text{ref}}\|_{\mathbf{W}_{\mathbf{u}}}^2, \\
    &l_{r,N}(\mathbf{x}_N) = \|\epsilon_{\mathbf{x}_N}\|_{\mathbf{W}_{\mathbf{x}_N}}^2 \nonumber,
\end{align}
where $\epsilon_{\mathbf{x}_i}:=\mathbf{x}_i\ominus\mathbf{x}_{i,\text{ref}}$ incorporates the $\mathrm{SO}(3)$ error, where the body rotational error is defined as $\epsilon_{\mathbf{R}_i}:=\text{log}(\mathbf{R}_{i,\text{ref}}^T\mathbf{R}_i)^{\vee}$ using the logarithm map and the $vee$ operator.\endnote{We use the inverse of right-Jacobian for exponential coordinates to get the Jacobian of $\epsilon_{\mathbf{R}_i}$ as described in~\citet{hong2020real}.} Errors in translation, joint configuration, and generalized velocity, are the differences between the desired and actual state. The corresponding diagonal weight matrices $\mathbf{W}_{\mathbf{x}}$, $\mathbf{W}_{\mathbf{u}}$, and $\mathbf{W}_{\mathbf{x}_N}$ are for states, control inputs, and the terminal state, respectively. 

\subsubsection{Reference Trajectory.} 
In all experiments and 3D simulations with the quadruped robot, we solely set the reference body's position and orientation in $\mathrm{SE}(3)$. All the other components are left at zero or in their fixed nominal configurations. Specifically, both $\mathbf{\dot{q}}_{i,\text{ref}}$ and $\mathbf{u}_{i,\text{ref}}$ are set to zero. The joint component of $\mathbf{q}_{i,\text{ref}}$ remains fixed as the nominal stance configuration; leg movements are regulated based on this nominal configuration, implying no pre-planned references for leg trajectories or foothold positions. We maintain the consistent $\mathbf{x}_{i,\text{ref}}$ without interpolating across all $i \in \{0,1,\cdots,N\}$, highlighting that fixed, elementary reference trajectories suffice for desired motions.

\subsubsection{Foot Slip and Clearance Cost.}
\begin{figure}
    \centering
    \includegraphics[width=1.0\columnwidth]{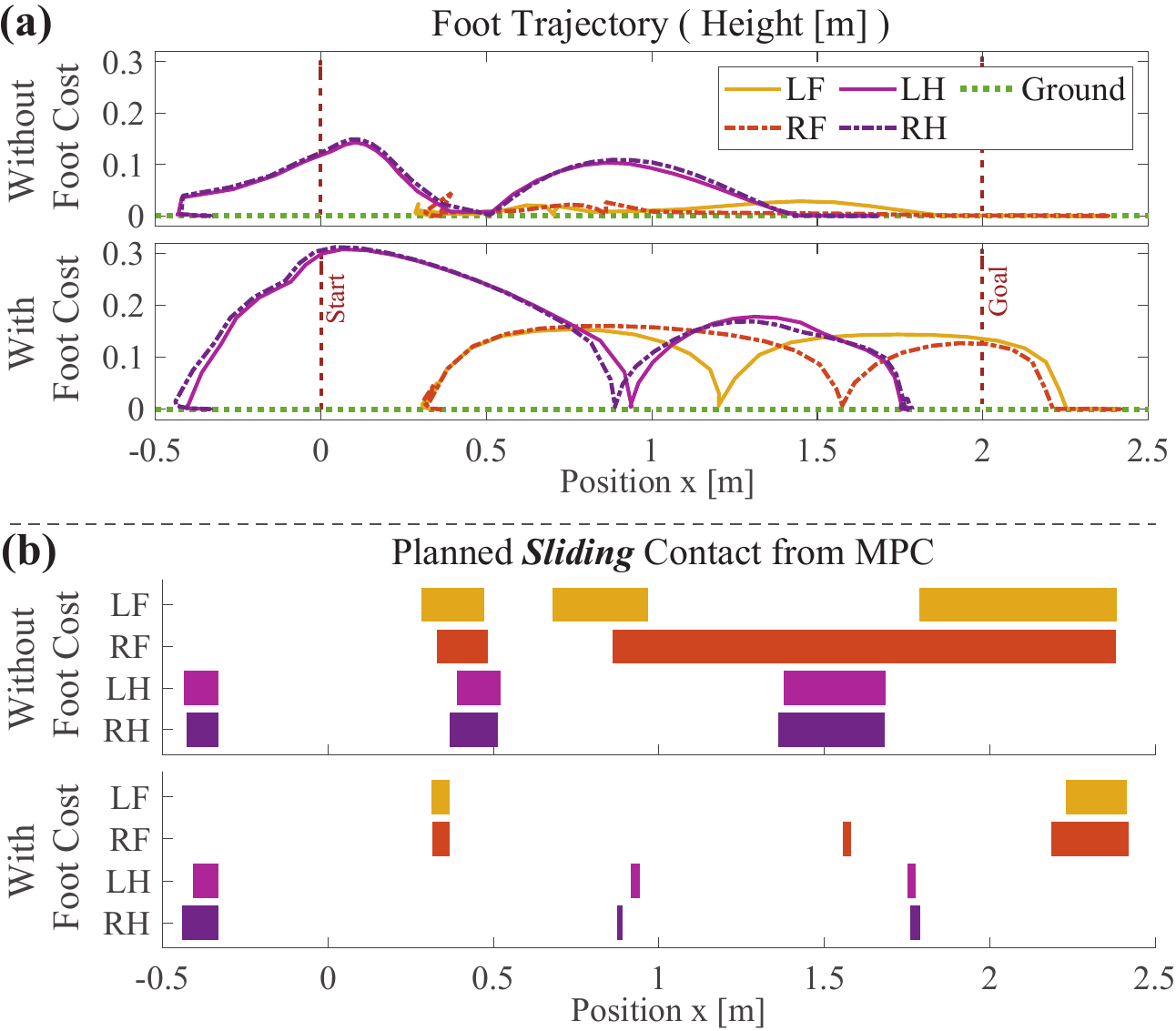}
    \caption[The effect of foot slip and clearance cost.]{ The effect of foot slip and clearance cost is depicted. Two scenarios are considered: "With Foot Cost", incorporating both regulating cost and foot slip and clearance cost, and "Without Foot Cost", using only the regulating cost. For both cases, the target position is set 2.0 m ahead of the initial position. The resultant swing leg trajectory is depicted in (a), and the planned sliding contact is illustrated in (b).}
    \label{figure:method_foot_slip_clearance_comparison}
\end{figure}

In the absence of pre-planned references for leg trajectories or foothold positions, solely using the regulating cost does not improve foot lifting or prevent slipping motion. This can result in minimal foot-lifting~\citep{Posa1} or prevalent scuffing~\citep{carius2019trajectory}, leading to challenges in real-world tracking and potential failures. Similar to~\cite{shirai2022simultaneous}, we add additional cost for these issues. Inspired by the foot slip cost and foot clearance cost commonly used in reinforcement learning framework~\citep{Hwangbo_2019,ji2022concurrent}, we formulate the individual costs as a single differentiable function for foot slip and clearance cost: 
\begin{align}
\label{eq:cost_foot}
    l_f(\mathbf{x}_i) = c_f\sum_{k=1}^{4} S(c_1\phi_{k,i})||\mathbf{v}_{k,i}^t||^2,
\end{align}
where $S(c_1\phi_{k,i})=1/(1+e^{(-c_1\phi_{k,i})})$ is the Sigmoid function to map the foot height as a scalar weight. $\phi_{k,i}$ and $\mathbf{v}_{k,i}^t$ denote the foot height and the tangential foot velocity, respectively. $c_1$ tunes the steepness of the Sigmoid function (set to $-30$), and $c_f$ is the total weight of the cost. Here, $S(c_1\phi_{k,i})$ approaches a value of 0.5 for small foot heights ($\phi_{k,i}\approx 0$), and approaches zero as the foot height $\phi_{k,i}$ increases. As a result, the tangential foot velocity is penalized when $\phi_{k,i}\approx 0$ (preventing foot slip), and the foot is encouraged to be high when the foot has tangential velocity (ensuring foot clearance), as shown in Figure~\ref{figure:method_foot_slip_clearance_comparison}.

\subsubsection{Air Time Cost.}
Considering various contact points, there could be multiple contact modes to track a reference configuration, allowing a quadruped to reach its target position using fewer than four legs, such as walking with only three legs. Specific gaits, such as trotting, often require fine-tuning of weights to balance body motion, joint motion, and control input~\citep{TO-w-contact}. In our work, we introduce an air time cost that penalizes prolonged swing leg time during walking, without the need for highly fine-tuned parameters for specific gaits:
\begin{align}
\label{eq:cost_airtime}
    & l_a(\mathbf{x}_i) = \sum_{k=1}^{4} {c_a}[k,i]\phi_{k,i}^2,
\end{align}
where ${c_a}[k,i]$ denotes the scalar weight of the air time cost for the $k$-th contact point at time step $i$. All \( {c_a}[k,i] \) values are initialized to 0, and are activated as a positive weight \( c_a \) only when swing leg time surpasses threshold \( i_t = 12 \) (\( 0.3~\text{sec} \)). From the initialized trajectory, we identify legs with excessive swing time over the entire horizon. For those legs, \( {c_a}[k,i] \) is set to \( c_a \) for time steps \( i_t \) to \( i_t+3 \) (during four time steps), penalizing foot lift with \( \phi_{k,i}^2 \). Once activated, this weight propagates in subsequent MPC problems until it reaches the beginning of the horizon (time step \( 0 \)).

\subsubsection{Symmetric Control Cost.}
The symmetric control cost aims to facilitate symmetric gait motions such as trotting and pacing. Here, the symmetry is emphasized within paired legs.\endnote{In contrast to previous method in \citet{yu2018learning}, which emphasizes global symmetry through mirroring actions of counterparts (e.g., right legs mirroring left legs), our cost focuses on within-pair symmetry (e.g., right legs together, left legs together).} The cost penalizes the torque differences for the hip flexion/extension (HFE) and knee flexion/extension (KFE) of the paired legs: 
\begin{align}
\label{eq:cost_symmetric}
    l_s(\mathbf{u}_i) = c_s||\mathbf{C}_2\mathbf{u}_i||^2,
\end{align}
where the $c_s$ denotes the scalar weight and $\mathbf{C}_2 \in \mathbb{R}^{4\times n_j}$ (with \( n_j = 12\)) is a constant matrix which maps the joint torques to the torque differences between the paired legs. In trot motion, diagonal legs are paired, while in pacing motion, same-side legs are paired. For example, in the case of diagonal pair, $\mathbf{C}_2^{\text{diagonal}}$ is:
\[
\mathbf{C}_2^{\text{diagonal}} = \begin{bmatrix}
\mathbf{D} & \mathbf{0}_{2 \times 3} & \mathbf{0}_{2 \times 3} & -\mathbf{D} \\
\mathbf{0}_{2 \times 3} & \mathbf{D} & -\mathbf{D} & \mathbf{0}_{2 \times 3} \\
\end{bmatrix},
\]
where the matrix \( \mathbf{D} \) is defined as $\mathbf{D} = [\mathbf{0}_{2 \times 1}, \mathbf{I}_{2 \times 2}]$.
Importantly, this cost function is not time-dependent, indicating that it is not activated periodically to induce a specific gait motion.

\subsection{Roll-out for the Forward Pass}
\label{sec:generation_rollout}
In the forward pass, the occurrence of contact is determined by evaluating the foot height condition, $\phi_{k,i}<0 \text{ or } \phi_{k,i+1}<0 ~\forall k \in \{1,\cdots,4\}$, based on the current state $\mathbf{x}_i$ and control $\mathbf{u}_i$. By approximating the next foot height using a first-order Taylor expansion, this condition ($\phi_{k,i+1}<0$) is represented as a velocity space condition with a drift term, thus leading to an affine condition for the contact impulse: 
\begin{align}
    \phi_{k,i+1}=\text{FK}_k(\mathbf{q}_{i+1})=\text{FK}_k(\mathbf{q}_i+\dot{\mathbf{q}}_{i+1}dt) \nonumber \\ 
    \approx \mathbf{J}_k^n(\mathbf{q}_i)\dot{\mathbf{q}}_{i+1}dt + \text{FK}_k(\mathbf{q}_i) = v^{n}_{k,i+1}dt+\phi_{k,i}, \label{eq:drift}
\end{align}
where $\text{FK}_k$ denotes the forward kinematics of the $k$-th contact point. Then, for the identified contact case, we compute the contact impulse based on the strict complementarity constraint, as detailed in Section~\ref{sec:contact_dyn}. For preventing the occurrence of penetration due to the large $dt$, we enforce $v^{n}_{k,i+1}+\frac{\phi_{k,i}}{dt} = 0$ instead of the original velocity space constraint $v^{n}_{k,i+1} = 0$. The details of this drift compensation are described in Appendix~\hyperref[sec:appendixB]{B}. Subsequently, the resultant contact impulse is used to integrate the state of the next time step via the semi-implicit Euler method.
 
\subsection{Gradient for Backward Pass}
In the backward pass, we compute the analytic gradient of the contact impulse based on the relaxed complementarity constraint. Then, the computed gradients of contact impulse are used in obtaining the gradient $\mathbf{f}_{\mathbf{x}}$ and  $\mathbf{f}_{\mathbf{u}}$ using the analytical derivative of the articulated body algorithm (ABA)~\citep{carpentier2018analytical,Pinocchio}. Notably, the entire gradient computation takes around $70~\mu s$ under the most demanding scenario with all four contact points in clamping cases, tested on a desktop PC with an AMD Ryzen 5 3600X processor. The details of the computation are in Appendix~\hyperref[sec:appendixB]{B}.

\section{Implementation Detail for Motion Execution}
\label{sec:implementation_motion_execution}
The optimization process is initialized by utilizing the state-control trajectory $\mathbf{x}_0, \ldots, \mathbf{x}_{N-1},$ $\mathbf{u}_0, \ldots, \mathbf{u}_{N-2}$, which is shifted by one time step from the solution of the previous MPC problem.\endnote{To ensure a safe initialization, we set the last control input $\mathbf{u}_{N-1}$ to zero, instead of directly copying $\mathbf{u}_{N-2}$ and subsequently simulate the last state $\mathbf{x}_{N}$. This mitigates potential issues, such as divergence in the last joint state resulting from the direct copy of incompatible large torque.} 
This initialization provides an effective starting point for subsequent optimization problem~\citep{diehl2002real}. It enhances convergence and guides the optimization towards the intended local minimum, ensuring motion continuity. 

However, the \textit{actual} initial state, represented by $\mathbf{\tilde{x}}_0$ (the current estimated state), often deviates from $\mathbf{x}_0$, the state anticipated in the previous MPC problem. 
This discrepancy leads to complications within contact-implicit approach.
In single shooting schemes, initialization is performed by successively integrating the control input trajectory, $\mathbf{u}_0, \ldots, \mathbf{u}_{N-1}$, starting from the \textit{deviated} initial state, $\mathbf{\tilde{x}}_0$, rather than $\mathbf{x}_0$. Given the hybrid model inherent in the contact-implicit method, even slight discrepancies can prompt a cascade of contact mode changes throughout the prediction horizon~\citep{Posa1}. This frequently results in trajectory divergence, leading to motion failures.

To address the aforementioned challenges, the FDDP algorithm \citep{mastalli2020crocoddyl,mastalli2022feasibility}, a multiple-shooting variant of DDP, is employed, as multiple shooting~\citep{bock1984multiple} could reduce the high sensitivity issue of single shooting~\citep{wensing2023optimization}. FDDP accepts state-control trajectories for initialization, where the initialized state trajectory, $\mathbf{x}_0, \ldots, \mathbf{x}_{N}$, implicitly guides contact modes consistent with the previous solution. Therefore, the initial state discrepancy ($\mathbf{\tilde{x}}_0 \neq \mathbf{x}_0$) does not lead to divergence of trajectory due to unwanted contact mode variations such as in single shooting scheme. Although the dynamics gap exists due to initial state discrepancy, the FDDP algorithm progressively reduces these gap in the early stage of optimization.

In the following subsections, we detail this issue by contrasting the single shooting and multiple shooting schemes in the contact-implicit MPC framework. Additionally, an overview of the feedback control is provided.

\begin{figure}
    \centering
    \includegraphics[width=1.0\columnwidth]{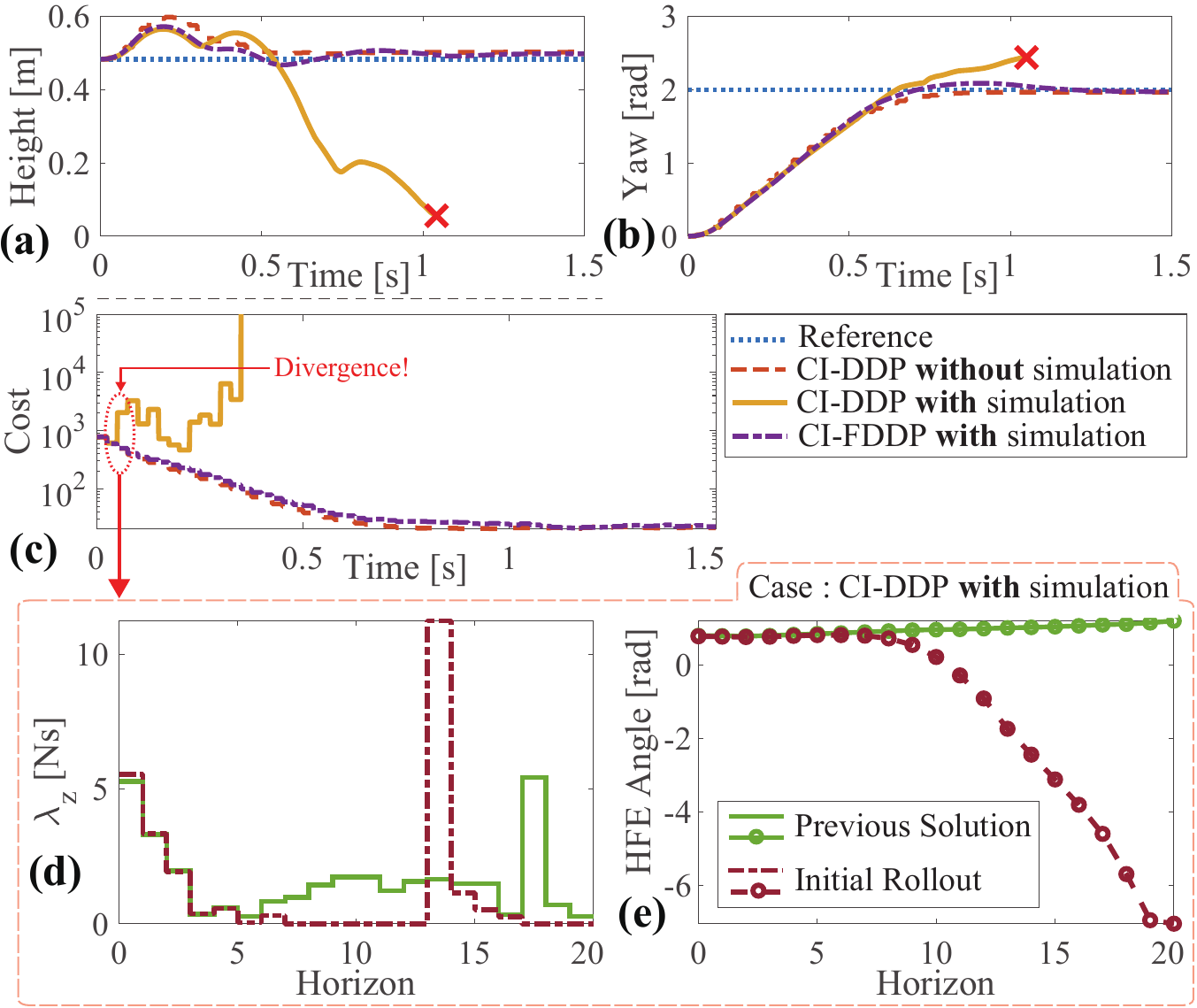}
    \caption[DDP initial divergence]{ Comparison of Contact-implicit MPC performance under the three cases: 1) CI-DDP \textbf{without} simulation. 2) CI-DDP \textbf{with} simulation. 3) CI-FDDP \textbf{with} simulation. (a) and (b) represent the resulting height and yaw, respectively, while (c) indicate the cost associated with each sequential MPC problem. In the ``CI-DDP \textbf{with} simulation" scenario, the red cross `x' marks in (a) and (b) denote failure due to body contact with the ground, while initial divergence is marked with red dots in (c). For this instance of divergence, the deviations of the state trajectory from the previous MPC solution and the current initial roll-out trajectory are depicted in (d)-(e): (d) RF leg's normal contact impulse, and (e) RF leg's HFE angle.}
    \label{figure:ddp_initial_divergence}
\end{figure}

\subsection{Single Shooting in Contact-implicit MPC}
\label{sec:execution_discrepancy}
The results in Figure~\ref{figure:ddp_initial_divergence} demonstrate the initialization issue of the single shooting scheme stemming from initial state discrepancies. The contact-implicit MPC performance is analyzed under the three cases: 1) CI-DDP \textbf{without} simulation. In this scenario, the initial state, $\mathbf{\tilde{x}}_0$, is directly taken as $\mathbf{x}_0$. This ensures that there is no initial state discrepancy, in line with solving a trajectory optimization problem in a receding horizon fashion. 2) CI-DDP \textbf{with} simulation. 3) Contact-implicit FDDP (CI-FDDP) \textbf{with} simulation. For 2) and 3) cases, RaiSim simulator~\citep{raisim} with 1 ms integration time step is used, and the MPC runs at 40 Hz. In these cases, the initial state discrepancies are inevitable, primarily due to the first-order discretization with a relatively large time step of $dt=25$ ms~\citep{Manchester2}. 
The reference body yaw angle is set as 2.0 rad, inducing a turning motion. 

In the case of ``CI-DDP \textbf{without} simulation'', the contact-implicit MPC using a single shooting method generates multi-contact motion, achieving the desired configuration, as depicted in Figure~\ref{figure:ddp_initial_divergence} (a)-(c). However, the case of ``CI-DDP \textbf{with} simulation'' exhibits divergence in the early stages of the overall motion, as shown in Figure~\ref{figure:ddp_initial_divergence}. 

In CI-DDP cases, initialization is achieved by a roll-out of the torque trajectory.
However, within the hard contact model, a slight height difference of foot can lead to unplanned contact loss. In Figure~\ref{figure:ddp_initial_divergence} (d), the right front (RF) leg is shown to lose ground contact prematurely. Instead of maintaining contact throughout the entire prediction horizon (N=20), the foot loses contact after just 5 time steps. Consequently, this alters the torque deployment, originally planned for ground reaction force (GRF), accelerating the joint velocity and eventually resulting in the divergence of joint position, observed in Figure~\ref{figure:ddp_initial_divergence} (e). Such changes also pose challenges in maintaining balance due to insufficient GRF. 
As a result, the initialized trajectory deviates significantly from the solution trajectory of the previous step. 
 

\subsection{Multiple Shooting for Contact-implicit MPC}
\label{sec:execution_multiple}

As evident in ``CI-FDDP \textbf{with} simulation'' of the yaw turning task depicted in Figure~\ref{figure:ddp_initial_divergence} (a)-(c), the CI-FDDP accomplishes the task despite initial state discrepancies. At the start of motion, the first MPC problem is initialized with a feasible state-control trajectory, outlined in following subsection. Then, the dynamics gap is exclusively attributed to the initial state discrepancy, and that dynamics gap at the 0 th iteration can be expressed as follows:
\begin{align}
    & \mathbf{\bar{f}}_0 := \mathbf{\tilde{x}}_0 \ominus \mathbf{x}_0 \neq \mathbf{0} \nonumber, \\ 
    & \mathbf{\bar{f}}_{i+1} := \mathbf{f}(\mathbf{x}_i,\mathbf{u}_i) \ominus \mathbf{x}_{i+1} = \mathbf{0}, \quad{\scriptstyle \forall i=\{0,1,...,N-1\}} \nonumber.
\end{align}
During the line search procedure in the forward pass, the initial roll-out state $\mathbf{\hat{x}}_0$ is determined along the step size $\alpha$ as follows: $\mathbf{\hat{x}}_0:=\mathbf{\tilde{x}}_0 \oplus (\alpha-1) \mathbf{\bar{f}}_0$, where it denotes the interpolation point lying between $\mathbf{\tilde{x}}_0$ and $\mathbf{x}_0$. If the roll-out trajectory is approved according to the acceptance condition, then the roll-out state $\mathbf{\hat{x}}_i$ is adopted as shooting state $\mathbf{x}_i$, and this entire process iterates until the gap is reduced to zero. 

\begin{figure}
    \centering
    \includegraphics[width=1.0\columnwidth]{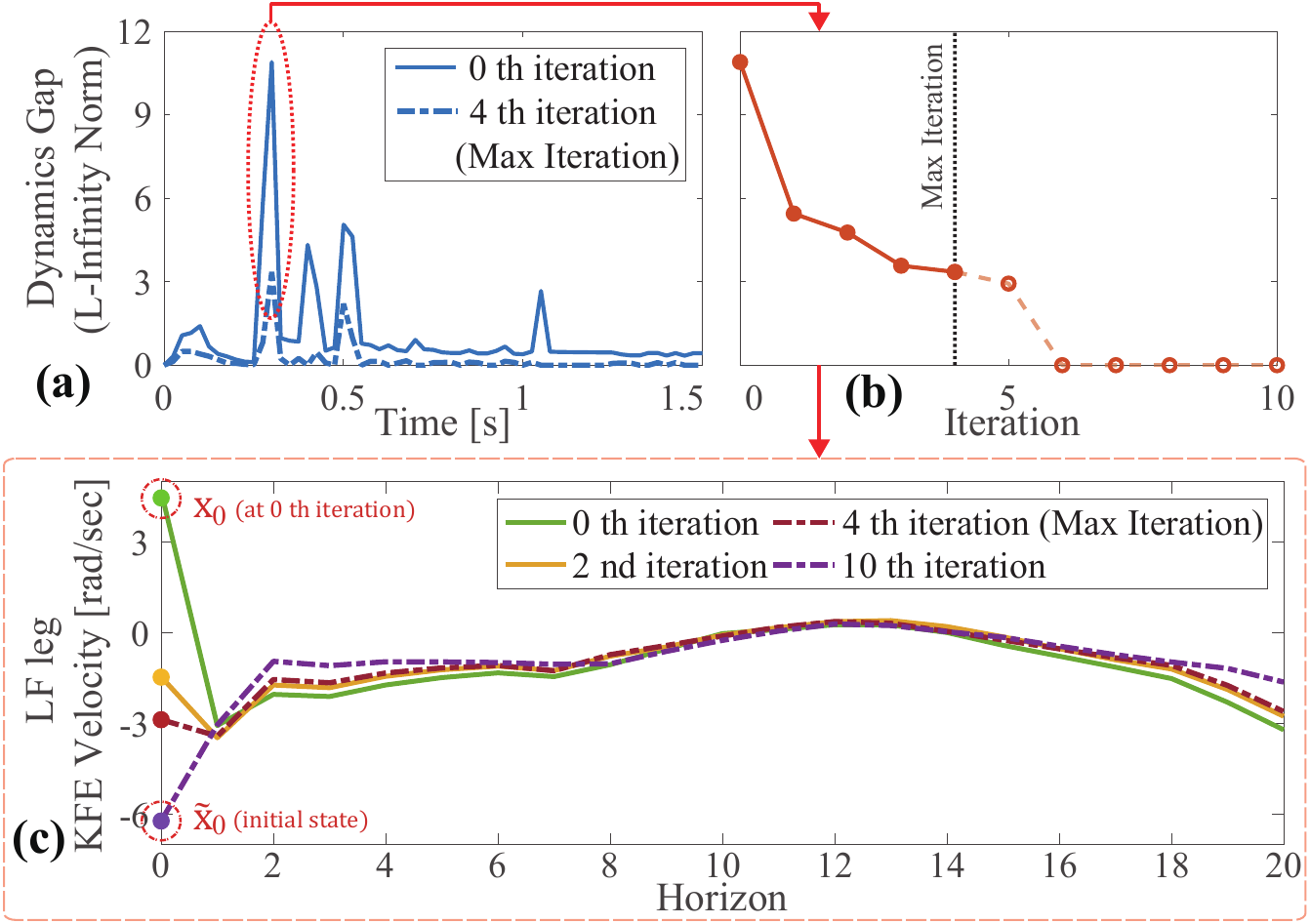}
    \caption[CI-FDDP]{The detailed description for the case of CI-FDDP for the yaw turning task, shown in Figure~\ref{figure:ddp_initial_divergence}. (a) denotes the dynamics gap at the initial and final iteration of MPC problems at each time step, with a maximum iteration set to 4. (b) illustrates the gap reduction process across iterations for the instance with the largest gap, marked by a red circle in (a). (c) denotes the LF leg KFE velocity trajectory along the horizon at each iteration.}
    \label{figure:cifddp_gap_contraction}
\end{figure}

A detailed process of gap contraction is presented in Figure~\ref{figure:cifddp_gap_contraction}. Figure~\ref{figure:cifddp_gap_contraction} (a) shows the dynamics gap at each MPC problem's initial and final iterations across the time steps. The detailed gap reduction process is illustrated in Figure~\ref{figure:cifddp_gap_contraction} (b). Furthermore, the shooting state trajectory is updated along the iterations, observed in Figure~\ref{figure:cifddp_gap_contraction} (c). In each iteration, the optimizer incrementally adjusts the trajectory by pulling the 0 th shooting state $\mathbf{x}_0$ closer to the initial state $\mathbf{\tilde{x}}_0$, facilitated by the progressive modification using step size $\alpha$, as shown in Figure~\ref{figure:cifddp_gap_contraction} (c). Meanwhile, the other states are closely preserved to the state of the previous iteration, due to the feedback term (gain K in equation~\eqref{eq:forward_FDDP}), implicitly guiding the contact modes to align with prior modes.

In conclusion, during the initialization, CI-FDDP prioritizes consistency with prior contact modes by trading off feasibility, thereby enhancing the robustness of initialization. Conversely, CI-DDP risks deviating from prior contact modes and often diverges as a result, when integrating the state trajectory to achieve feasible initialization. In practice, CI-FDDP also manages unexpected contact events (Section~\ref{result_slip_recovery}), and conducts real-robot experiments despite model imperfections and state estimation uncertainties.

\subsubsection{Initial Feasible Trajectory.}
\label{sec:cifddp_initialization}
Prior to the initial MPC problem, we solve a trajectory optimization problem under the reference configuration with a standing posture, using the same framework. This aims to initialize the initial MPC with a \textit{feasible} state-control trajectory. All varied motions in this paper stem from this standing motion initialization, not requiring the task-specified pre-planned trajectories.

\subsubsection{Maximum Iteration.}
\label{sec:execution_maxiteration}
To ensure real-time computation requirements, we limit the maximum number of iterations of the CI-FDDP as 4. The optimizer requires minimal iterations to reduce the dynamics gap and explore contact modes in each iteration. As evident in Figure~\ref{figure:cifddp_gap_contraction} (a), despite the optimizer's inability to eliminate all gaps in every instance due to the limitation of maximum iteration, the subsequent MPC problems could compensate it with fast re-computation setting.

\subsection{Feedback Control}
\label{sec:execution_feedback}
Similar to \citet{youm2023imitating}, we proposed a PD control that combines feed-forward torques with feedback actions: 
\begin{align}
    & \mathbf{p}_{\text{target}} = \mathbf{u}^{*}_{0}/K_p + \mathbf{p}_{\text{des}} \nonumber, \\ 
    & \mathbf{u}_{\text{cmd}} = \mathbf{K}_p(\mathbf{p}_{\text{target}}-\mathbf{p}) + \mathbf{K}_d(\dot{\mathbf{p}}_{\text{des}}-\dot{\mathbf{p}}) \nonumber.
\end{align}
Here, $\mathbf{p}_{\text{des}}$ and $\dot{\mathbf{p}}_{\text{des}}$ are chosen based on the next time step's joint state from the solution trajectory of the CI-FDDP. If the current joint position and velocity, $\mathbf{p}$ and $\dot{\mathbf{p}}$, are close to $\mathbf{p}_{\text{des}}$ and $\dot{\mathbf{p}}_{\text{des}}$, respectively, $\mathbf{u}_{\text{cmd}}$ aligns with the feed-forward torque $\mathbf{u}^{*}_{0}$. Feedback actions increase as they deviate from $\mathbf{p}_{\text{des}}$ and $\dot{\mathbf{p}}_{\text{des}}$.

In our target scenario of dynamic motions, such as front-leg rearing, frequent contact mismatches occur due to rapid changes in contact mode. PD control could robustly handle these unexpected contact events. For example, in the late (or missing) contact cases, the PD control could decelerate an accelerated leg, and in the early contact cases, it could put more torques for the additional ground reaction forces. In our work, this straightforward implementation of PD control is sufficient to execute the motions planned by the MPC \endnote{Also, in many contact-implicit approaches~\citep{TO-w-contact,neunert2018whole,carius2018trajectory,carius2019trajectory,aydinoglu2023consensus}, a PD control is utilized to address the issues from missed contact. Feedback gains from iLQR are another option, but in our implementation, they did not outperform PD control. iLQR gains work well in the vicinity of the planned trajectory. However, in our contact-implicit framework with hybrid dynamics, slight state deviations can significantly change control effects due to contact state changes (moving beyond the vicinity).}.

\section{Results}
\label{sec:results}

We present the results of our framework using quadruped robots. First, we analyze the effect of the relaxation variable $\rho$ using a 2D planar model. Next, we showcase multi-contact motions for various desired configurations in 3D simulations, and compare our framework with MuJoCo MPC~\citep{howell2022predictive}. Lastly, we validate the contact-implicit MPC framework with hardware experiments, illustrating trot and front-leg rearing motions.

\begin{figure}
    \centering
    \includegraphics[width=1.0\columnwidth]{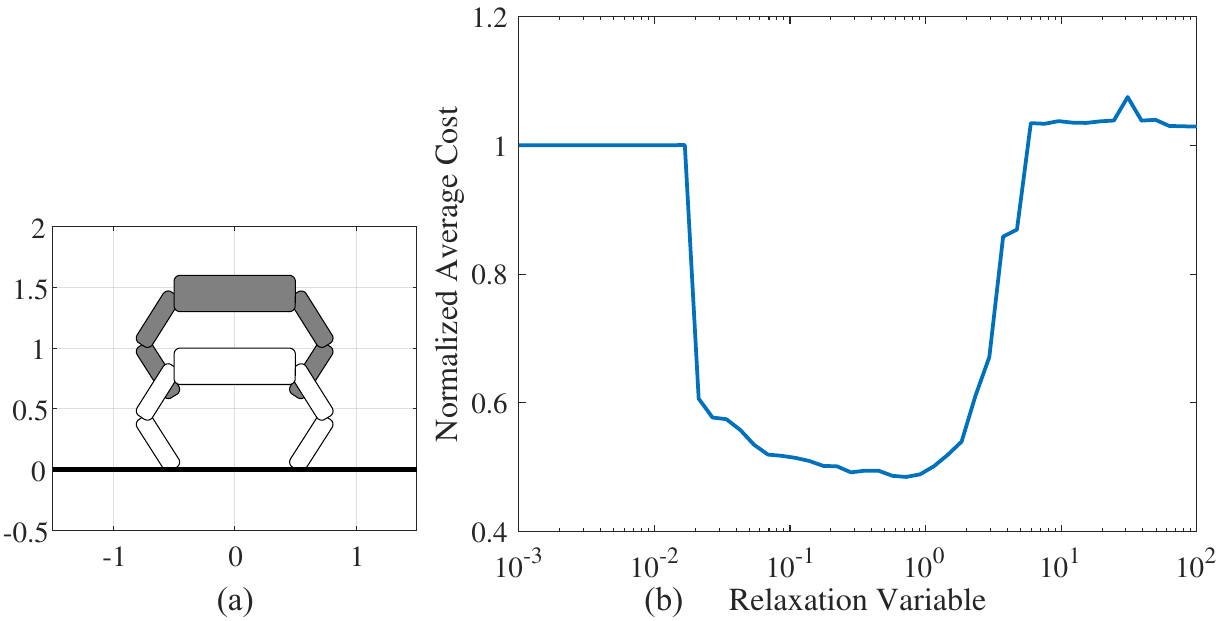}
    \caption[Relation between relaxation variable and cost for jumping up task]{Relation between relaxation variable and average cost for jumping up task : (a) 2D environment, grey robot for the reference trajectory and white robot for actual robot. (b) Result of the average cost according to the relaxation variable.}
    \label{figure:relaxationVariableCost}
\end{figure}

\subsection{Effect of the Relaxation Variable}
\label{effectRelaxation}
We investigate the impact of the relaxation variable $\rho$, which allows the gradient to navigate motions that involve breaking contact. We conducted a 2D simulation with a planar quadruped robot, following the settings from our earlier work~\citep{kim2022contact}. The framework solely employed a regulating cost with a fixed reference configuration. Figure~\ref{figure:relaxationVariableCost} (a) illustrates the 2D setup and Figure~\ref{figure:relaxationVariableCost} (b) maps the average cost against the relaxation variable.

\begin{figure}
    \centering
    \includegraphics[width=1.0\columnwidth]{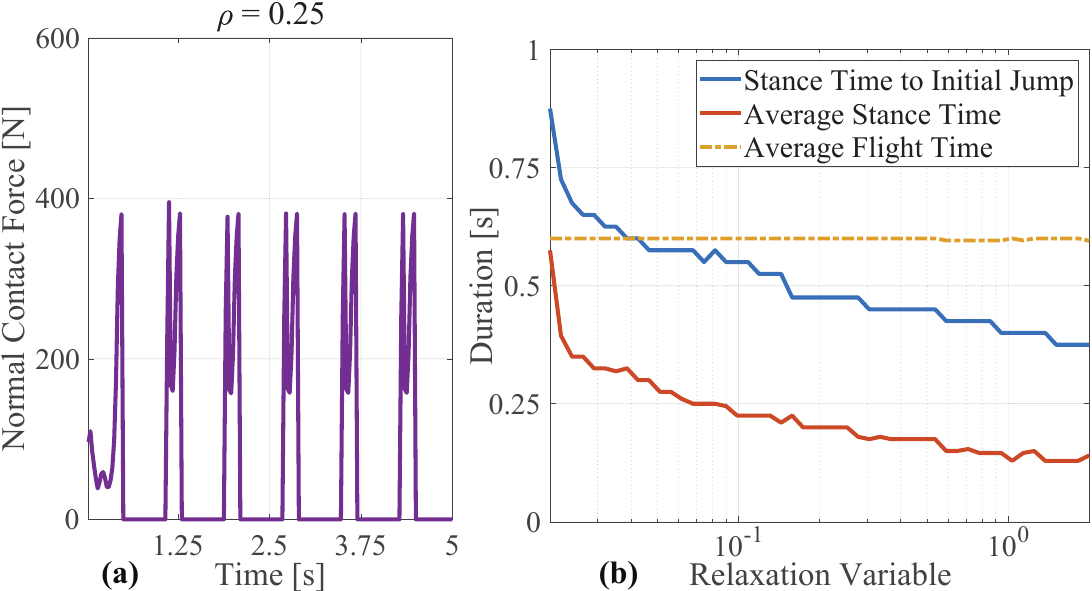}
    \caption[Relation between relaxation variable and the stance time for jumping up task]{Periodic jumping motion analysis: (a) Contact impulse for 0.25 relaxation variable, indicating periodic jumps. (b) Average stance time versus relaxation variable.}
    \label{figure:relaxationVariable}
\end{figure}

In this scenario, with a reference trajectory set significantly above the robot, periodic jumping results in a lower cost than standing still. Without relaxation, the motion remains standing, aligning with the initialized trajectory where all feet are in contact, until the relaxation variable exceeds 0.02. Beyond this value, the robot exhibits a periodic jumping motion, achieving a lower cost than standing. However, when the relaxation variable exceeds 6, the robot prematurely attempts to jump without accumulating sufficient contact impulse. This leads to in-place bouncing, increasing the cost due to unnecessary movement, as illustrated in Figure~\ref{figure:relaxationVariableCost} (b). All the discovered motions are in Extension 3. 

Notably, there are no modifications to the cost function, such as weight adjustments or time-dependent costs. The smooth gradient (through relaxation) enables the emergence of the jumping motion, leading to a lower-cost solution within the same optimal control problem.

\begin{figure}
    \centering
    \includegraphics[width=1.0\columnwidth]{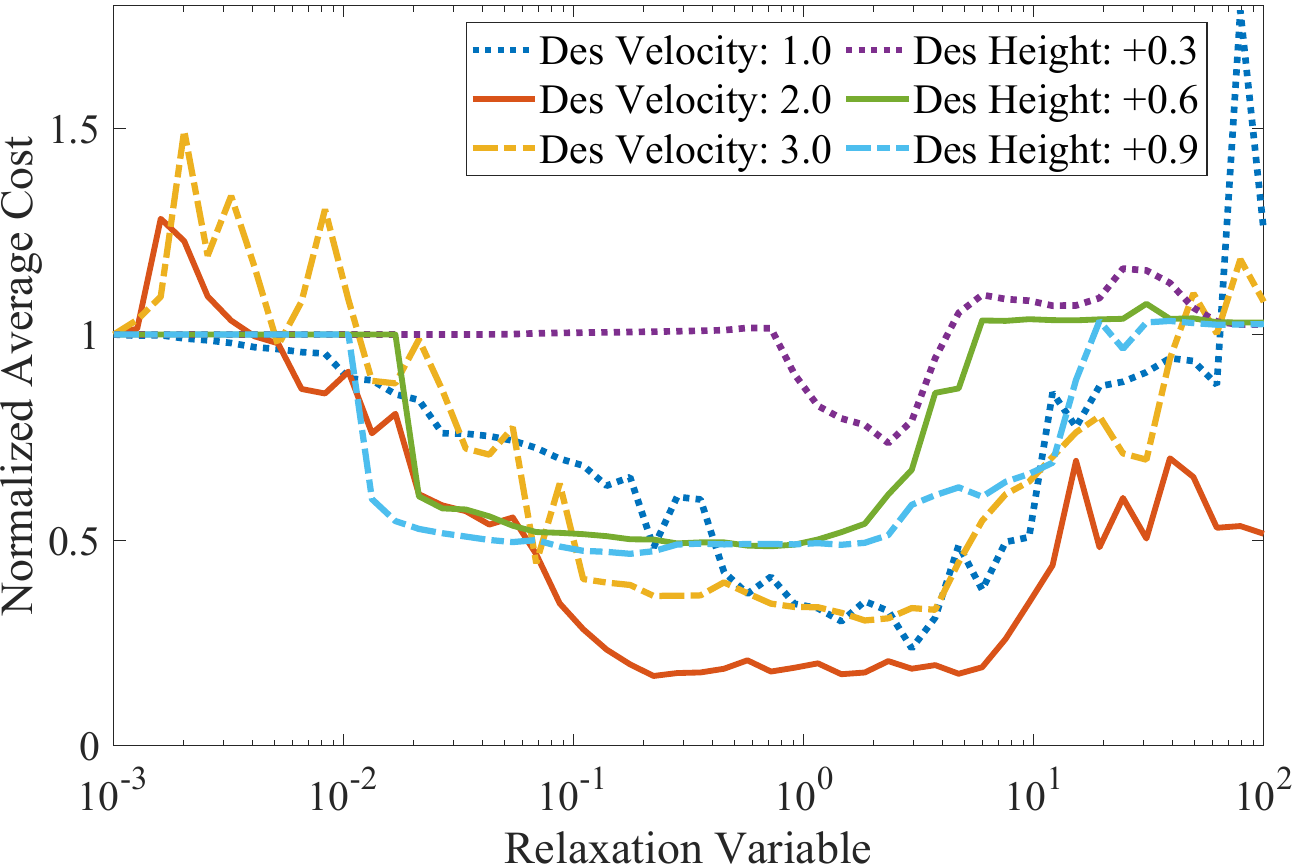}
    \caption[Relation between relaxation variable and cost for various task]{ Result of the cost according to the relaxation variable for jumping up task 0.3~ m, 0.6~ m, and 0.9~ m, and moving forward task 1.0~ m/s, 2.0~ m/s, and 3.0~ m/s.}
    \label{figure:relaxationGuitar}
\end{figure}

We further investigate the effect of relaxation variable $\rho$ within the scope of [0.02, 2], exhibiting periodic jumping motions. Figure~\ref{figure:relaxationVariable} (a) depicts the contact impulses for jumping, and Figure~\ref{figure:relaxationVariable} (b) presents the average stance time. Notably, there is a consistent decrease in average stance time as $\rho$ increases (Figure~\ref{figure:relaxationVariable} (b)), implying that the $\rho$ guides the optimizer to initiate contact-breaking motions. 

\begin{figure*}
    \centering
    \includegraphics[width=2.0\columnwidth]{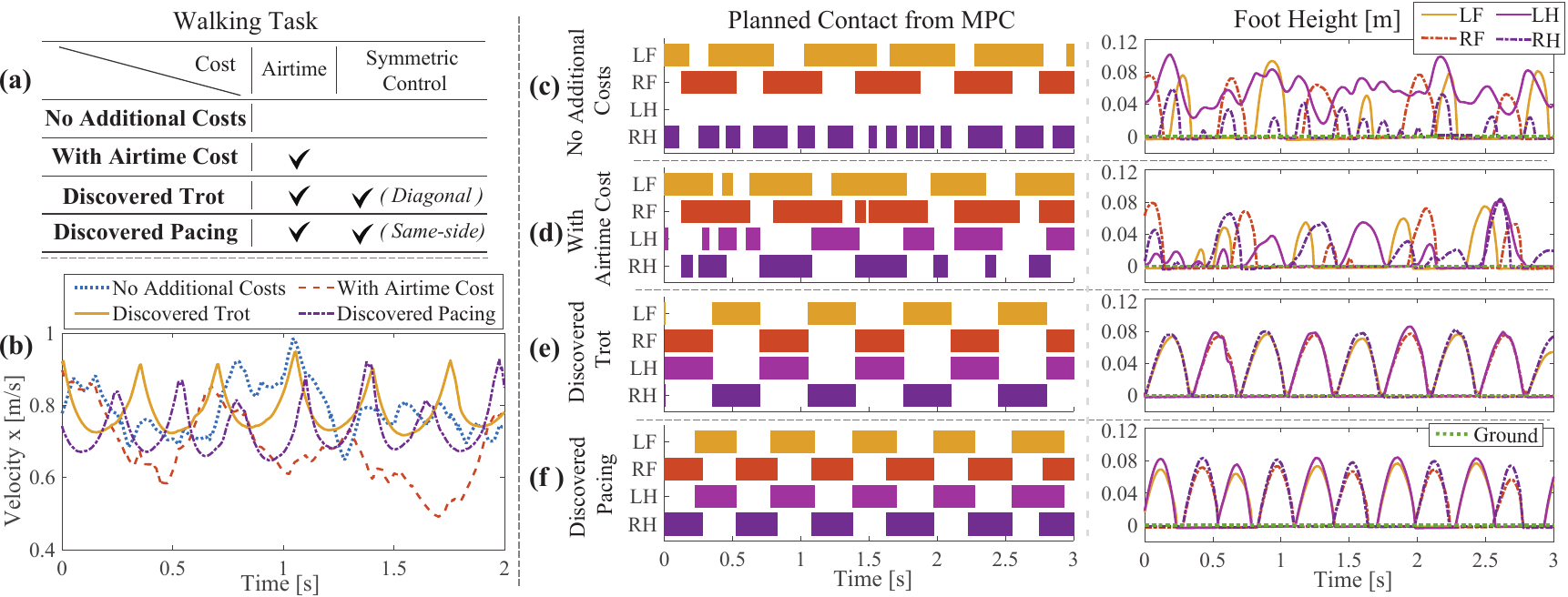}
    \caption[Simulation results of walking task along the costs.]{Analysis for the walking motions under different cost configurations: (a) Table showing cost components for each case. (b) Velocity profiles for each walking motion. (c)-(f) Display planned contact and swing leg trajectory on the z-axis.}
    \label{figure:Simulation_walking}
\end{figure*}

Achieving a lower cost utilizing a smooth gradient is evident across various tasks. Figure~\ref{figure:relaxationGuitar} presents the average cost according to the relaxation variable for both `jumping up' and `moving forward' tasks. In the jumping-up task, reference trajectories are placed 0.3 m, 0.6 m, and 0.9 m above the robot's starting position, remaining constant. For the moving forward task, reference trajectories maintain a consistent forward velocity of 1.0 m/s, 2.0 m/s, and 3.0 m/s. As shown in Figure~\ref{figure:relaxationGuitar}, across different tasks, a consistent region around [0.1, 10] emerges where the cost is reduced below that of the no-relaxation case, implying that the effect of the smooth gradient is not task-dependent.

Therefore, the smooth gradient can facilitate the emergence of diverse motions as shown in Extension 8, including forward movement, continuous flipping, and sliding, all of which are achieved with a simple reference trajectory in 2D environment.


\subsection{3D Simulation Results}
\subsubsection{Implementation of 3D Environment.}
The HOUND quadruped robot was employed~\citep{2022_Shin_HOUND}. We fixed the relaxation variable $\rho$ at a value of $2.0$ for HOUND without additional adjustments for all simulations. Detailed weight settings can be found in Appendix~\hyperref[sec:appendixB]{B}. The entire 3D simulation process was facilitated by the RaiSim simulator~\citep{raisim}, with a 1 ms integration time step, devoid of control delays. For each MPC instance, the exact simulated state was taken as the initial state. 

\subsubsection{Various Quadruped Motions.}
To validate the capacity of our contact-implicit MPC in generating diverse multi-contact motions, we display the results in Extension 4. The reference configuration is formed by modifying only the body reference position and orientation in $\mathrm{SE}(3)$, including: (a) Target pitch: 0.6 rad, (b) Target pitch: 1.2 rad, (c) Target roll: 0.6 rad, (d) Target height: nominal height + 0.25 m. In the Extension 4, the transparent robot represents the reference configuration, while the colored spheres on the ground indicate expected contact points planned in the current MPC problem. All motions are derived primarily using the regulating cost and foot slip and clearance cost.

We showcase both cases: \textit{with relaxation} (proposed, $\rho=2$) and \textit{no-relaxation} ($\rho=0$). Without the smooth gradient (\textit{no-relaxation}), diverse multi-contact motions do not manifest under the same conditions. Instead, the resulting motions converge to a tilted posture with all feet in contact.

\begin{figure*}
    \centering
    \includegraphics[width=2.0\columnwidth]{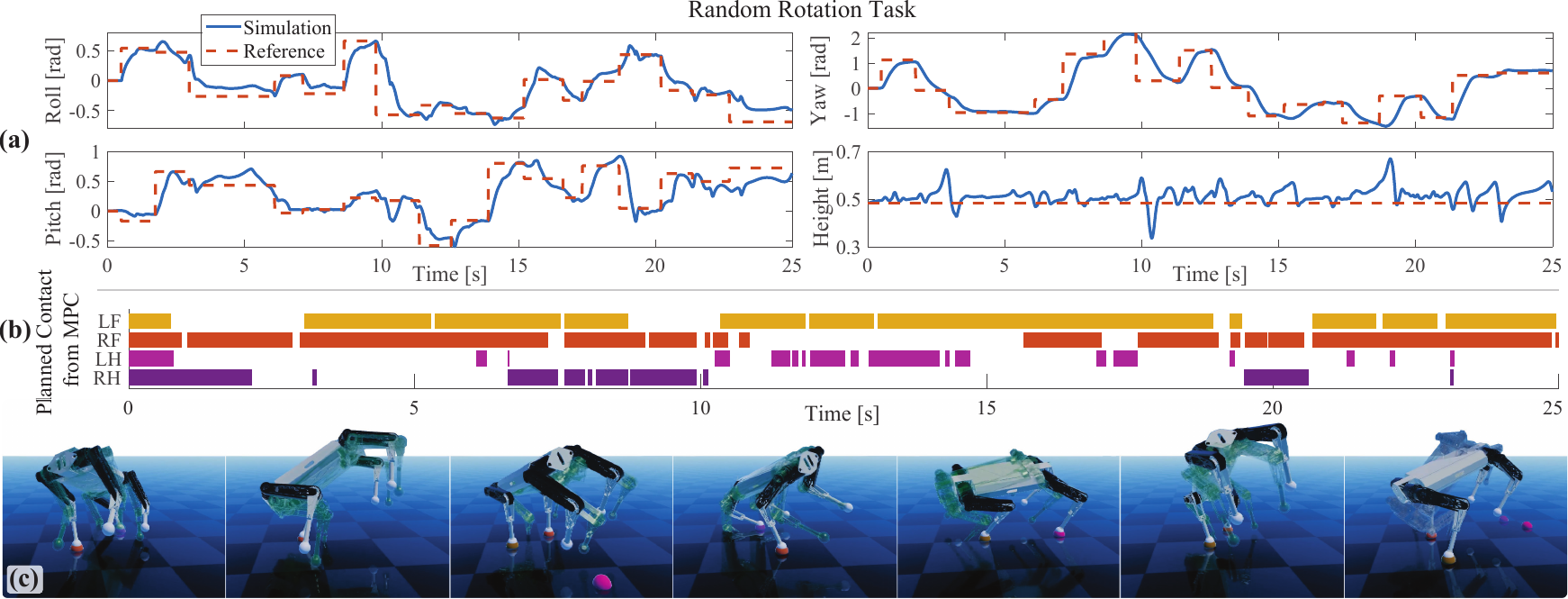}
    \caption[Simulation results of random rotation task.]{Analysis for the random rotation task, when the robot aligns with the reference configuration, the reference is subsequently changed: (a) Desired and resulting rotation angle (roll, pitch, yaw) and height. (b) MPC-planned contact. (c) Snapshots, where the green transparent robot indicates the alignment condition being satisfied.}
    \label{figure:Simulation_random_rot}
\end{figure*}

With the proposed method, front-leg rearing motions emerge from a desired pitch reference. The quadruped tries to balance on its front leg by adjusting foothold positions or swiftly kicking its hind leg to leap upwards, aiming to match the desired configuration. It is noticeable that the desired configuration does not specifically guide the CoM positioning beneath the support line between the front legs. Even when the desired posture is intrinsically unstable and not the system's equilibrium, our method aims to align with it repeatedly (Extension 4). This feature is also emphasized in the random rotational task (Figure~\ref{figure:Simulation_random_rot}, and Extension 6).

Likewise, a side-leg rearing motion occurs with the desired roll angle reference, and a jumping motion occurs with the desired height offset. The snapshots of the resulting motions are exemplified in Appendix~\hyperref[sec:appendixB]{B}.

\subsubsection{Discovery of Gait.}
\label{result_walking}
We demonstrate that our framework identifies a walking motion without a periodic gait-encoded cost (e.g., bending knee periodically)~\citep{neunert2018whole,carius2019trajectory,kim2022contact}. Instead, we utilize air time cost (equation~\eqref{eq:cost_airtime}) and symmetric control cost (equation~\eqref{eq:cost_symmetric}). The outcomes are illustrated in Figure~\ref{figure:Simulation_walking} and Extension 5. At each MPC problem, the reference configuration is set to be 0.55 m ahead of the current robot. 


Utilizing only basic costs (regulating, foot slip and clearance), a three-leg walking motion emerges, as shown in Figure~\ref{figure:Simulation_walking} (c). The robot does not utilize the left hind (LH) leg and relies on the right hind (RH) leg, leading to chattering in the RH foot due to insufficient swing time. Introducing an air time cost yielded a four-legged motion, as in Figure~\ref{figure:Simulation_walking} (d), engaging the LH leg and reducing the load on the RH leg. However, the motion remains non-cyclic with a bias towards the front legs bearing most of the GRF, and a significant velocity deviation (Figure~\ref{figure:Simulation_walking} (b)).


Incorporating a symmetric control cost, our framework discerns gait motions for trotting and pacing, as shown in Figure~\ref{figure:Simulation_walking} (e) and (f), respectively. Despite all the costs not being periodically activated and the short 0.5-sec horizon, the resultant gait is cyclic and remains consistent across sequential MPC problems. Additionally, the optimizer naturally selects distinct frequencies for trotting and pacing gaits. Also, the load required to support the body is evenly distributed through the front and hind legs, and the smooth swing leg trajectory is observed (Extension 5). 

\subsubsection{Random Rotational Task.}
\label{result_random_rot}
In this subsection, we explore the random rotation task. The target rotation is randomly set and adjusted upon meeting the success criteria. This task showcases our framework's ability to handle various contact combinations in response to quickly changing reference configurations. The reference body rotation is uniformly sampled as roll, pitch $\sim U(-0.8, 0.8)$ rad, and yaw difference $\sim U(-2,2)$ rad, with the yaw angle updated accordingly. The success criteria for the reference body rotation, \( \mathbf{R}_{\text{ref}} \), is met when the angle of the rotation difference is under \(10\,~\text{deg}\), \( \|\epsilon_{\mathbf{R}}\|<0.174\,~\text{rad} \) (with \( \epsilon_{\mathbf{R}} \) computed as \( \text{log}(\mathbf{R}_{\text{ref}}^T\mathbf{R}_{\text{current}})^{\vee} \)), and the body's lateral deviation is below 0.4 m. The motion trajectory and reference body rotation are depicted in Figure~\ref{figure:Simulation_random_rot} (a), while the identified contact sequences are displayed in Figure~\ref{figure:Simulation_random_rot} (b). Figure~\ref{figure:Simulation_random_rot} (c) displays the motion snapshots. Dynamic motions emerge, such as single and double-leg balancing and turning, rearing maneuvers, pivoting turns around a leg, and swift kick-back, which are available in Extension 6. 

\subsubsection{Recovery from Unexpected Contact Events.}
\label{result_slip_recovery}

\begin{figure*}
    \centering
    \includegraphics[width=2.0\columnwidth]{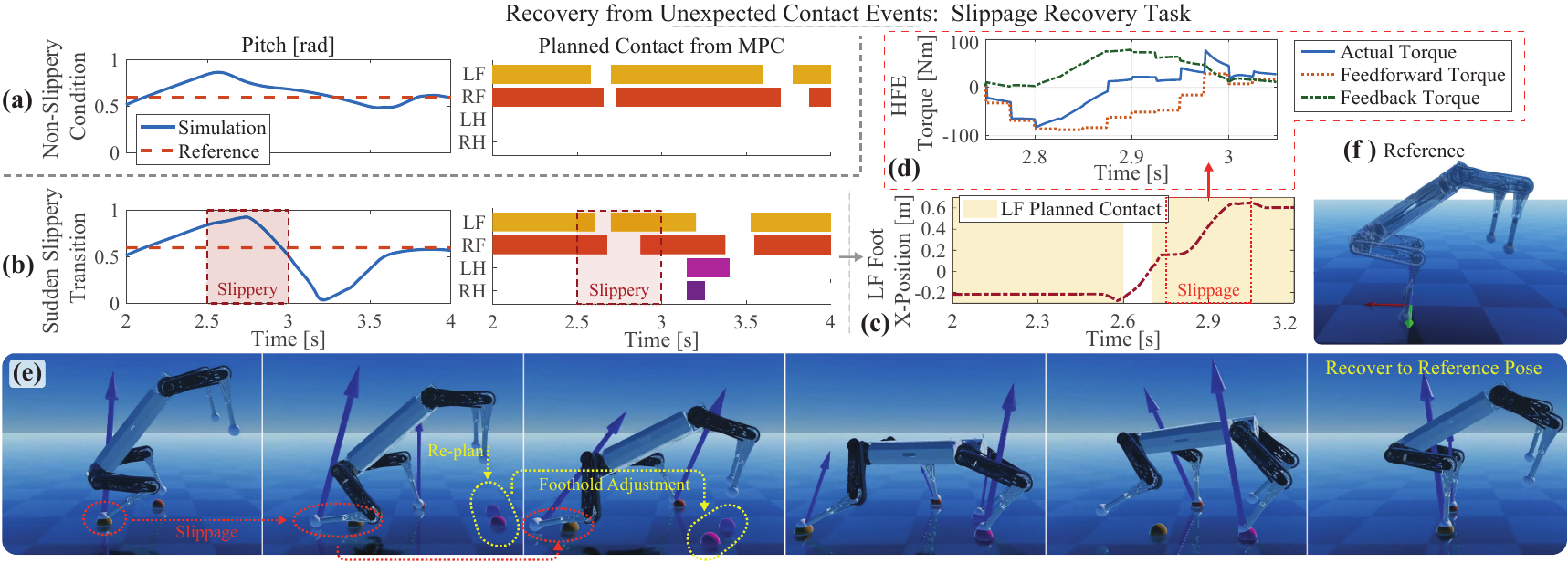}
    \caption[Simulation results of slippage recovery task.]{
    Analysis of the slippage recovery task, which involves an unexpected transition to slippery terrain, resulting in slippage: (a) Without transition to slippery terrain, (b) With transition to slippery terrain. Both scenarios (a) and (b) illustrate the reference and actual pitch angles, along with MPC-planned contacts. Subfigures (c)-(e) detail an instance of slippage involving the left front (LF) leg and the subsequent recovery behavior: (c) LF foot x-axis position, (d) LF leg's HFE torque profiles including feedforward and feedback components, (e) Snapshots demonstrating re-planning and foothold adjustments after the slippage, with colored spheres indicating the planned next contact points. (f) Reference configuration.}
    \label{figure:Simulation_slip_recovery}
\end{figure*}

The re-planning capabilities of the proposed framework are demonstrated through artificially induced unexpected contact events: 1) Slippage Recovery Task, 2) Missed Contact Recovery Task. In the first scenario, unexpected slippage is induced by a sudden change in the ground's friction coefficient (from 0.8 to 0.25 and back to 0.8). In the second scenario, missed contacts occur, such as no contact (even when planned), early contact, or late contact, due to a sudden change in the ground's height (from 0 to -0.1 m or 0 to 0.1 m). In both cases, the changes are not reflected in the MPC model and remain unknown. Nevertheless, the robot swiftly recovers through the feedback policy and by re-planning in subsequent MPC problems. We present the recovery motions across various reference configurations in Extension 9 for slippage and Extension 10 for missed contact, without further tuning on weight. 

For clarity, Figure~\ref{figure:Simulation_slip_recovery} illustrates the slippage recovery process with a reference pitch angle of 0.6 rad. Under non-slippery conditions, a front-leg rearing motion emerges, as shown in Figure~\ref{figure:Simulation_slip_recovery} (a). However, with a sudden, unknown friction drop (from 2.5 sec to 3.0 sec), slippage of the left front (LF) foot occurs (Figure~\ref{figure:Simulation_slip_recovery} (c)), causing the robot to fail to balance on only the front two feet (Figure~\ref{figure:Simulation_slip_recovery} (b)). Initially, the PD control addresses the LF leg's positional deviation, by exerting torque in the opposite direction of the slippage (Figure~\ref{figure:Simulation_slip_recovery} (d)). Subsequent MPC problems aim to stabilize the body by adjusting the other legs, notably re-planning to engage two hind legs and shifting the planned foot hold positions (Figure~\ref{figure:Simulation_slip_recovery} (e)). In particular, following the slippage, the hind legs support the falling body, and the robot executes a backward kick to return to the reference pose (Figure~\ref{figure:Simulation_slip_recovery} (e) and Extension 9). Despite unexpected disturbances, the framework swiftly adapts, recovers, and continues with the task’s progression.

\subsubsection{Comparison with MuJoCo MPC.}
We benchmark our framework against MuJoCo MPC~\citep{howell2022predictive,Yuval}, a contact-implicit MPC framework. For the comparison, the iterative linear quadratic gaussian (iLQG) method in the MuJoCo MPC library is used, leveraging MuJoCo's finite-difference gradient computation. Both frameworks tackle the identical optimal control problem. Specifically, a regulating cost is utilized with identical weights: generalized coordinate (pos xy, pos z, rotation, joint pos) set as (1, 10, 10, 0.1), generalized velocity (joint vel) at 5e-4, and control input at 5e-4. Both methods adopt a discretization time step of 20 ms for a 20-step horizon, totally to 0.4 seconds. The Unitree A1 quadruped robot is employed under the same control input constraints. 

\begin{figure}
    \includegraphics[width=1.0\columnwidth]{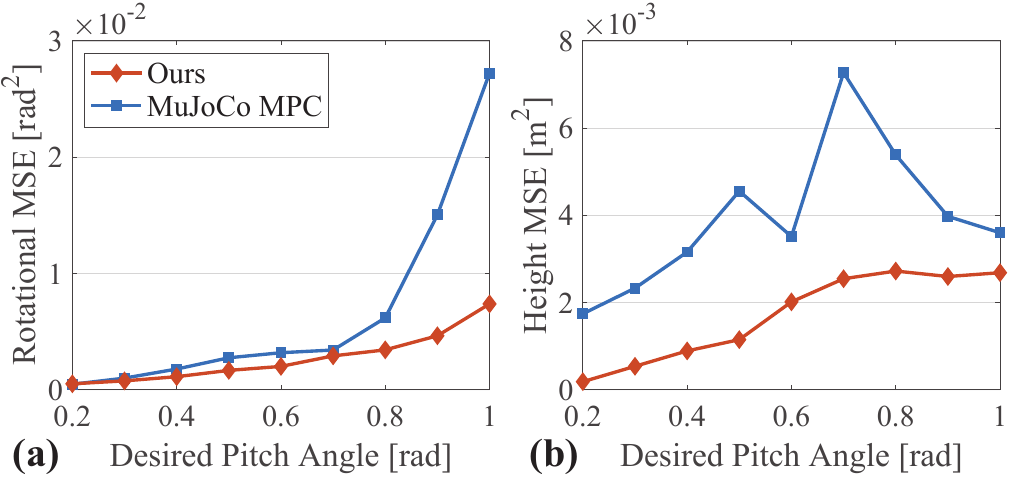}
    \caption[Comparison with MuJoCo MPC]{Comparing our proposed framework with the iLQG of the MuJoCo MPC for a desired pitch ranging from 0.2 rad to 1.0 rad, where each task spans 10 seconds: (a) Mean squared error (MSE) for rotational angle error. (b) MSE for height error.}
    \label{figure:Comparison_MuJoCoMPC}
\end{figure}

Our framework is tested on the Raisim simulator, while MuJoCo MPC employs the MuJoCo simulator~\citep{todorov2012mujoco}, both on the same desktop PC with an AMD Ryzen 5 3600X processor. The desired pitch angles are set from 0.2 rad to 1.0 rad, with a height goal of 0.28 m. Figure~\ref{figure:Comparison_MuJoCoMPC} presents a comparison of tracking errors.

\begin{figure}
    \centering
    \includegraphics[width=1.0\columnwidth]{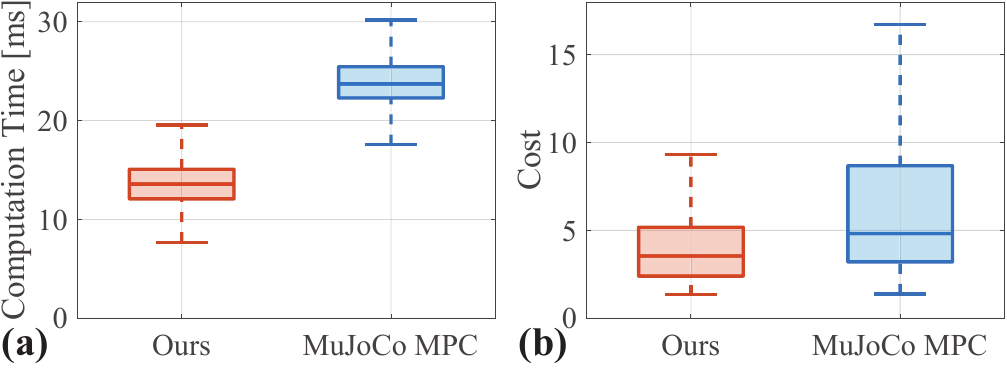}
    \caption[Comparison of computation time and Cost with MuJoCo MPC]{Box plot comparison of (a) computation times and (b) costs for all tasks in Figure~\ref{figure:Comparison_MuJoCoMPC}. Each task comprises 500 MPC problems, corresponding to a 10-second duration.}
    \label{figure:Computationtime_Cost_MuJoCoMPC}
\end{figure}

In our approach, the relaxation variable $\rho$ is set as 1.0. This explicit relaxation in the analytical gradient promotes repeated attempts at a rearing motion, thereby efficiently aligning the robot with both the desired pitch angle and height. This results in minimal angle error, with only slight compromises in height tracking for steeper pitch angles (0.7 rad to 1.0 rad). For the MuJoCo MPC approach, they employ finite differences combined with soft contact simulation~\citep{todorov2011convex,todorov2014convex}. This implies that the range of \textit{smoothness} could be limited due to the trade-off between \textit{smoothness} and contact feasibility of the soft contact model. The restricted smoothness can induce minimal contact breakage motions, such as body tilting with all foot contacts or handstanding with body touch (Extension 7). This can compromise height tracking (Figure~\ref{figure:Comparison_MuJoCoMPC} (b) for 0.5-0.8 rad desired pitch), and angle tracking (Figure~\ref{figure:Comparison_MuJoCoMPC} (a) for desired pitches over 0.8 rad). Therefore, our method generally results in a lower cost, as reflected in the overall cost distribution in Figure~\ref{figure:Computationtime_Cost_MuJoCoMPC} (b). Simulation results are in Extension 7.

The average computation time for each optimal control problem is shown in Figure~\ref{figure:Computationtime_Cost_MuJoCoMPC} (a). 
Given the real-time constraint (sampling time 20 ms), the maximum iteration for the problem is determined. 
Due to this constraint, the iLQG of the MuJoCo MPC operates in a single iteration, while our method allows up to four iterations.
These iterations in our approach aim to reduce the feasibility gap and explore alternative contact modes. 
Notably, even with the increased number of iterations, we achieve a shorter computation time, as depicted in Figure~\ref{figure:Computationtime_Cost_MuJoCoMPC} (a). 
This is attributed to our efficient analytical gradient computation.


\subsubsection{Comparison with Relaxed Forward Pass.}
\label{sec:comparison_relaxed}
We compare our framework against the relaxed forward pass approach, where both the forward and backward passes are based on contact model with relaxed complementarity constraints. Therefore, the relaxed forward model allows forces to act at a distance. 

\begin{figure}
    \centering
    \includegraphics[width=1.0\columnwidth]{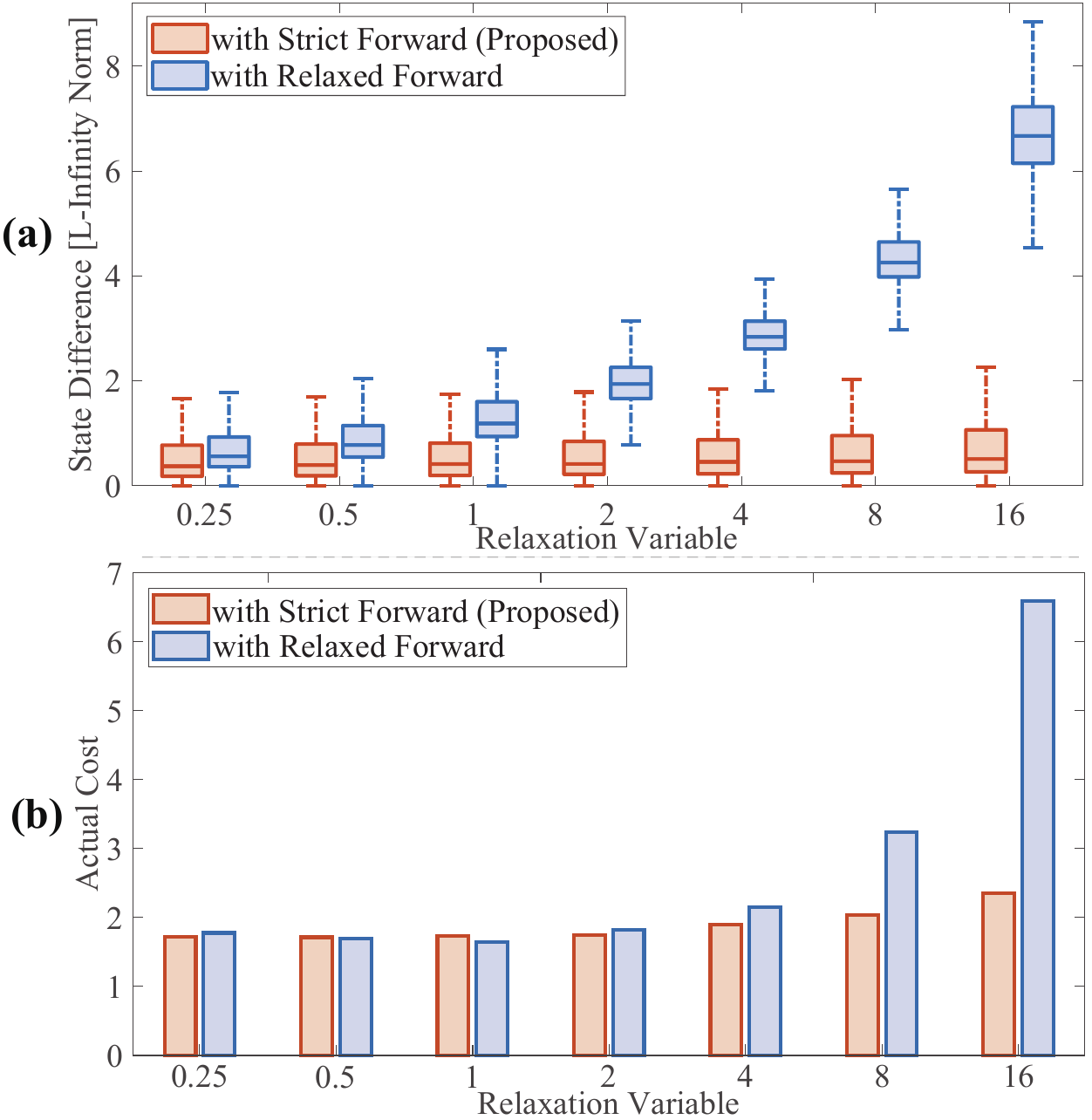}
    \caption[Comparison with relaxed forward pass]{Comparison with relaxed forward model: (a) State difference between the controlled and planned states (from MPC), and (b) actual cost computed at each time step (25 ms) based on the controlled state and control input.}
    \label{figure:Comparison_Relaxed_Forward}
\end{figure}

We analyzed 1) deviations between the planned (from MPC) and actual controlled states in Figure~\ref{figure:Comparison_Relaxed_Forward} (a), and 2) actual cost value based on the controlled state and control input in Figure~\ref{figure:Comparison_Relaxed_Forward} (b), for 50 randomly sampled tasks with reference body postures: roll $\sim U(-0.6, 0.6)$, pitch $\sim U(-0.8, 0.8)$, yaw $\sim U(-2.0, 2.0)$, x, y $\sim U(-0.8, 0.8)$, z offset $\sim U(-0.1, 0.1)$. Relaxation variables range from 0.25 to 16, centered around 2, which is used in our experimental setting. For our proposed method, the same relaxation variable is applied to the backward pass.

In the relaxed forward model, increasing the relaxation value amplifies non-physical contact effect, resulting in larger deviations between the controlled and planned states, as in Figure~\ref{figure:Comparison_Relaxed_Forward} (a). Consequently, the degradation of closed-loop performance, indicated by increased cost with relaxation variables exceeding 4, is shown in Figure~\ref{figure:Comparison_Relaxed_Forward} (b). These results stem from MPC plans based on unrealizable contact impulses. For example, the robot attempts to balance using non-physical contact forces (acting at a distance), repeatedly failing to make these contacts, even without disturbances. Reducing the relaxation value can mitigate this loss of physical realism, but compromises smoothness for optimization, as described in the previous section. 

Conversely, our framework avoids non-physical forces, ensuring achievable motions, as demonstrated by smaller state deviations and relatively consistent costs (Figure~\ref{figure:Comparison_Relaxed_Forward} (a) and (b), respectively). This motion fidelity is maintained even with increased relaxation in the backward pass, as the forward roll-out uses a strict model. Therefore, we could achieve both physical realism and smoothness for optimization.

\subsection{Experiments}
Finally, we conducted experiments to validate the proposed framework in two scenarios: 1) Executing a front-leg rearing motion with a fixed pitch reference configuration and 2) Performing trot motion directed by a joystick command.

\begin{figure*}
    \includegraphics[width=2.0\columnwidth]{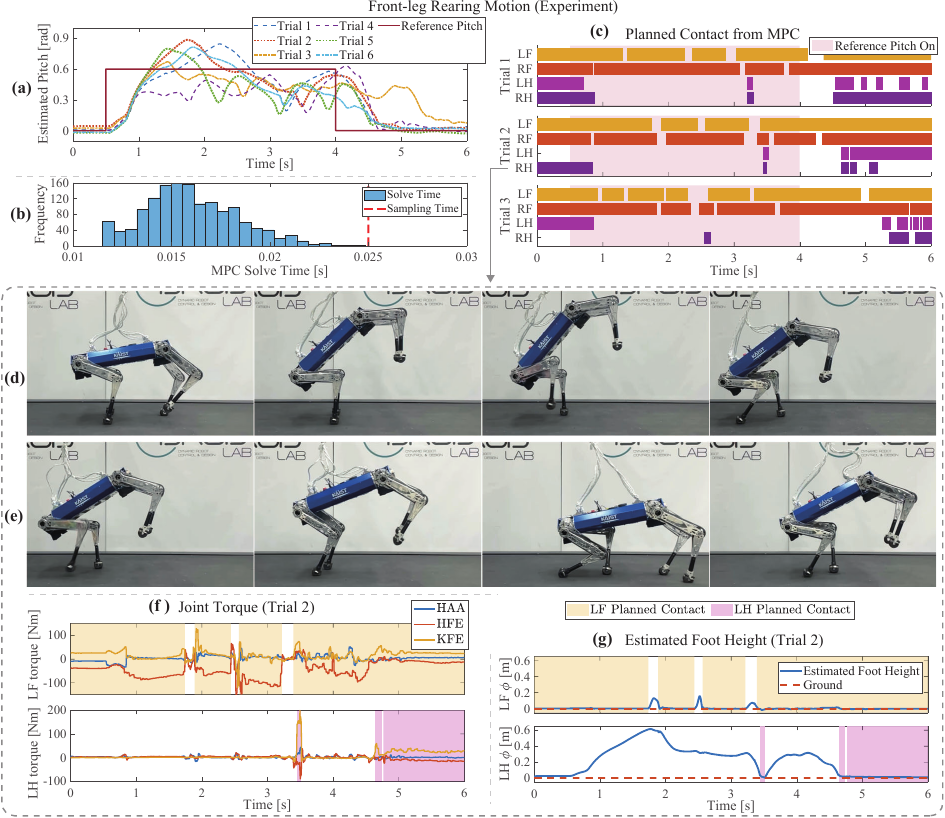}
    \caption[Experiment result of handstanding]{Experimental results of the front-leg rearing motion for 6 trials: (a) Estimated pitch angle with a target of 0.6 rad over 3.5 sec. (b) Histogram of MPC computation times. (c) MPC-derived contact sequences for Trials 1-3. (d-e) Motion snapshots, (f) Joint torques, and (g) Estimated foot heights for Trial 2.}
    \label{figure:Experiment_handstanding}
\end{figure*}

\subsubsection{Experimental Setup.}
The HOUND quadruped robot weighing 45 kg~\citep{2022_Shin_HOUND} was employed for the experiments. The maximum joint torques are 200 Nm for the hip abduction/adduction (HAA) and hip flexion/extension (HFE) joints and 396.7 Nm for the knee flexion/extension (KFE) joint. The maximum joint velocities are 19.2 rad/s for HAA and HFE joints and 9.7 rad/s for the KFE joint at 73 Volts (based on experimental findings). A single onboard computer with an Intel(R) Core(TM) i7-11700T CPU @ up to 4.6 GHz is utilized for solving the contact-implicit MPC problem, EtherCAT communication, PD control, and state estimation. The MPC problems are solved using Box-FDDP~\citep{mastalli2022feasibility}, consistent with the simulation environment, incorporating feedback PD control. The MPC runs at 40 Hz. The joint position and velocity are sampled at 2 kHz, leading the PD control to function at the same 2 kHz frequency, while the state estimator works at 1 kHz. 

\subsubsection{State Estimation.}
For body state estimation, we employ a linear Kalman filter algorithm that fuses leg kinematics and IMU information. This algorithm is used to estimate the body's linear position and velocity with respect to the global (world) frame. Additionally, we utilize a generalized momentum method~\citep{de2006collision} to estimate the foot contact state. This detected contact state informs the Kalman filter.\endnote{The detected contact state is also used to mitigate body height drift~\citep{bledt2018cheetah} by assuming that the height of supporting feet are zero.
}

To address the delay caused by the computation time of the MPC, we employ state prediction approach similar to \citet{mastalli2022agile}. We utilize the predicted generalized coordinate, $\mathbf{q}_{\delta}$, for the initial state $\mathbf{\tilde{x}}_0$ of the MPC problem. The prediction step is expressed as:
\begin{align}
    \mathbf{q}_{\delta}=\mathbf{q}+\dot{\mathbf{q}}\delta t \nonumber,
\end{align}
where $\mathbf{q}$ and $\dot{\mathbf{q}}$ represent the currently estimated generalized coordinate and generalized velocity, respectively. The time interval $\delta t$ is set to 18 ms, accounting for the average computation time and allowing for potential peak durations.

\begin{figure*}
    \includegraphics[width=2.0\columnwidth]{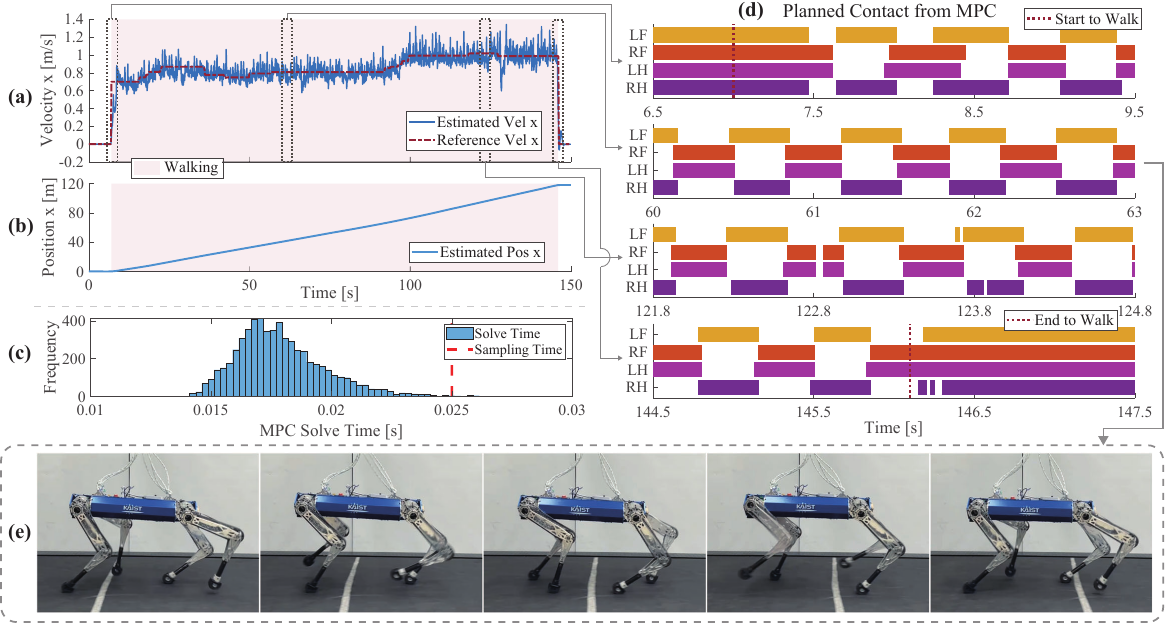}
    \caption[Experiment result of walking (trot)]{Experimental results of the walking motion for the discovered trot motion: (a) Estimated velocity with a target velocity in the x-direction. (b) Estimated x-position. (c) Histogram of MPC computation times. (d) MPC-derived contact sequences. (e) Motion snapshot.}
    \label{figure:Experiment_walking_trot}
\end{figure*}

\subsubsection{Front-leg Rearing Motion.}
\label{result_exp_rearing}
The reference is set to a pitch angle of 0.6 rad for making rearing motion, with a 0.3 m positional x offset and 0.1 m height offset. This desired configuration is maintained for 3.5 seconds, and then returning to a nominal configuration with a 0 rad pitch angle. The relaxation variable is set as 2.0, and the regulation cost and the foot slip and clearance cost are used with weights consistent with the simulation setting. The result of a total of 6 trials\endnote{The motion is tried 3 times consecutively with short-term breaks (under 5 seconds), and the experiment is conducted twice.} is depicted in Figure~\ref{figure:Experiment_handstanding} (a), and the capabilities of real-time computation is verified in Figure~\ref{figure:Experiment_handstanding} (b). For the first three trials, the planned contact sequences are shown in Figure~\ref{figure:Experiment_handstanding} (c). The torque trajectories and foot trajectories are shown in Figure~\ref{figure:Experiment_handstanding} (f) and (g), respectively. All trials are available in Extension 1. 

All trials exhibit varied motions, even though they show a common motion trend, as observed in Figure~\ref{figure:Experiment_handstanding} (a) and (c). This variation arises because the framework does not rely on a fixed trajectory, typically computed in offline trajectory optimization, but instead generates motions online in a real-time fashion. Consequently, it allows adaptive responses to each specific situation. 

The rearing motion starts with a hind leg kick, and the robot aims to stabilize its body pitch near the target angle using front feet. If the body leans forward, the robot either pivots on one front leg or adjusts both front feet to gain reverse momentum from the ground reaction forces, as shown in Figure~\ref{figure:Experiment_handstanding} (d). If leaning backward, the robot initially tries to balance using its front feet, and if the momentum continues downward, it resorts to a backward kick to elevate the body, depicted in Figure~\ref{figure:Experiment_handstanding} (e). In cases where the initial kick is not sufficient, the robot will attempt another kickback (Trial 4). Despite balancing failures or missed contacts, the robot adjusts using other legs, as demonstrated in Trials 3 and 4. In contrast to more reactive controllers that rapidly adjust to maintain balance, our MPC exhibits a forward-planning motion style, such as waiting for the hind legs to make contact with the ground before executing a kick-back, ensuring consistency across subsequent MPC problems.

\subsubsection{Trot Motion.}
To assess the viability of the real-time discovery of gait in subsection~\ref{result_walking}, we tested for the trot motion. Consistent with the simulation settings, there is no periodic cost applied to enforce a specific gait. Forward and yaw velocity commands, $v_{\text{des}}$ and $\omega_{\text{des}}$ respectively, dictate the reference configuration. This is computed as $(v_{\text{des}}+0.1)dtN$ in front of the robot's current pose, with a yaw deviation of $\omega_{\text{des}}dtN$. The results illustrate around 120 m treadmill walk along 140 seconds, as seen in Figure~\ref{figure:Experiment_walking_trot} (b). The velocity ranged between [0.7,1.0] m/s (Figure~\ref{figure:Experiment_walking_trot} (a)). This experiment allowed for online gait identification, revealing a trot as visualized in Figure~\ref{figure:Experiment_walking_trot} (d) and a snapshot in Figure~\ref{figure:Experiment_walking_trot} (e). MPC problem computation times are presented in Figure~\ref{figure:Experiment_walking_trot} (c). While no periodic cost was introduced to enforce a trot gait, the discovered gait showed minor deviations from a conventional trot at higher desired velocities, as observed in Figure~\ref{figure:Experiment_walking_trot} (d). Additionally, smooth gait transitions during the start and stop phases were also identified, as depicted in Figure~\ref{figure:Experiment_walking_trot} (d). The full video is available in Extension 2.

\subsubsection{Limitation of Experiment.}
While simulations allow for dynamic motion discovery, real-world experiments present difficulties due to the absence of an accurate state estimator.
Our framework depends on roll-out trajectories for contact planning, being sensitive to the precision of initial state estimations.
Although it can address certain discrepancies between expected and actual initial states (e.g., due to large discretization time step), the challenge here is that the framework plans based on an \textit{incorrectly estimated} state.
Specifically, body height estimation is crucial as it dictates potential contact modes by determining foot height through kinematics.
Inaccurate height estimation can lead to misplanned contacts and eventual failures of motion.
This becomes more critical in dynamic motions with fewer contacts, where balance relies solely on one or two feet\endnote{The effect of state estimation is tested in simulations using motions such as steep rearing or repeated jumping, where all four feet are often in the air, making estimation unreliable. Even with increased iterations for improved MPC convergence, the success rate does not improve. Incorrect state estimation inaccurately sets the optimization problem’s initial state, deviating from the robot’s true state. Thus, even with a fully reduced dynamics gap, control is based on an incorrect state, not the true initial state, failing to enhance the success of tasks where height estimation is unstable.}. 
Hence, while our framework handles a range of dynamic motions, extreme motions in experiments, such as rearing motions with longer duration or steeper desired pitch angles, remain an area for future work.

\section{Discussion}
In this section, we discuss the implications and potential limitations of the proposed method, focusing on the relaxed gradient and dynamics gap.
\label{sec:discussion}
\subsection{Gradient with Relaxation Variable.}
The relaxation variable introduces bias to the exact gradient to explore new contact modes and potentially achieve lower-cost solutions. While our results show potential benefits, there are limitations when the relaxed gradient deviates significantly from the exact gradient. Excessively large relaxation values can cause the gradient to deviate from the descent direction, resulting in line search failure to reduce costs. Consequently, convergence to local minima may not be achieved. However, in our simulations and experiments, we observed that this impact on convergence is slight within the tested range of relaxation variables. Consequently, while proper tuning of the relaxation variable is crucial, the process is straightforward. The fast computational speed allows efficient trial-and-error tuning, with simulations taking considerably short time (such as under 5 seconds for our system) for each trial. Details on convergence issues with excessive relaxation are provided in Appendix~\hyperref[sec:appendixC]{C}.

\subsection{Analysis about the Dynamics Gap}
The dynamics gap in the proposed method occurs when the initial state diverges from the planned state of the MPC model. It means that in scenarios involving relatively static motions with minimal deviations, the dynamics gap typically reduces to zero within the iterations. However, in dynamic motions such as front-leg rearing, a single contact mismatch can create a significant dynamics gap, which cannot be fully reduced within the maximum iteration setting. Although the gap may not close in the current MPC problem, subsequent MPC problems gradually close it, stabilizing the robot. Detailed analysis of the dynamics gap in the front-leg rearing experiment is presented in Appendix~\hyperref[sec:appendixC]{C}. 





\section{Conclusion}
\label{sec:conclusion}
In this work, we present a contact-implicit MPC framework that simultaneously generates and executes multi-contact motions without pre-planned contact mode sequences or
trajectories. The hard contact model based on the complementarity constraint is utilized in the DDP-based algorithm, with the analytic gradient of contact impulse. To explore new contact modes, especially involving breaking contact, an analytically computed smooth gradient of contact impulse is proposed, based on the relaxed complementarity constraint. By leveraging the smooth gradient, the framework can discover diverse multi-contact motions, starting from fully-contacted standing poses. To enhance the tractability of the motions such as enough foot lifting, we have incorporated differentiable cost terms. Furthermore, for ensuring stable motion execution, the multiple shooting variant of DDP is employed. Our framework's efficacy is demonstrated through various multi-contact motions using the 45 kg quadruped robot, HOUND, in both simulated environments and real-world experimental tests.

In this study, we mainly consider contacts between the feet of quadrupeds and ground to ease the computations. While a range of motions is identified with this consideration, a more generalized approach for future work would involve accounting for contacts between other body parts and various objects. 


Lastly, while our proposed method demonstrates potential, its sensitivity to state estimation has posed challenges in realizing highly dynamic motions in real-world experiments.
Future work could benefit from leveraging learning-based approaches to enhance robustness, such as employing learning-based estimation or integrating an auxiliary control term in place of the conventional PD control. Also, the diverse set of motions generated by our framework can serve as valuable prior motion data for learning-based controllers, as seen in works such as~\citet{peng2021amp,peng2022ase,wu2023learning}.

\begin{funding}
The author(s) disclosed receipt of the following financial support for the research, authorship, and/or publication of this article: This research was financially supported by the Institute of Civil Military Technology Cooperation funded by the Defense Acquisition Program Administration and Ministry of Trade, Industry and Energy of Korean government under grant No. UM22207RD2.
\end{funding}
\theendnotes


\clearpage
\appendix
\section*{Appendix A}
\subsection*{Index to Multimedia Extension}
\begin{table}[htbp]
\centering
\newcolumntype{C}[1]{>{\centering\arraybackslash}p{#1}}
\begin{tabular}{C{0.15\columnwidth} C{0.1\columnwidth} p{0.6\columnwidth}}
\toprule
\textbf{Extension} & \textbf{Media Type} & \textbf{Description} \\
\midrule
1 & Video & Real-world experimental results on a quadruped demonstrating front-leg rearing motion.\\
2 & Video & Real-world experimental results on a quadruped showcasing real-time discovery of a trot gait. \\
3 & Video & Resultant motion of varying the relaxation variable in a 2D environment. \\
4 & Video & 3D simulation results of various reference configurations. \\
5 & Video & 3D simulation results demonstrating gait discovery. \\
6 & Video & 3D simulation results for a random rotation task. \\
7 & Video & 3D simulation comparison between our method and MuJoCo MPC.  \\
8 & Video & Additional motion results in a 2D environment. \\
9 & Video & 3D simulation results of re-planning in response to slippage. \\
10 & Video & 3D simulation results of re-planning in response to missed contact events. \\
\bottomrule
\end{tabular}
\end{table}

\section*{Appendix B: Implementation Details}
\label{sec:appendixB}

\subsection*{Drift Compensation in Forward Pass}

\begin{figure}
    \centering
    \includegraphics[width=1.0\columnwidth]{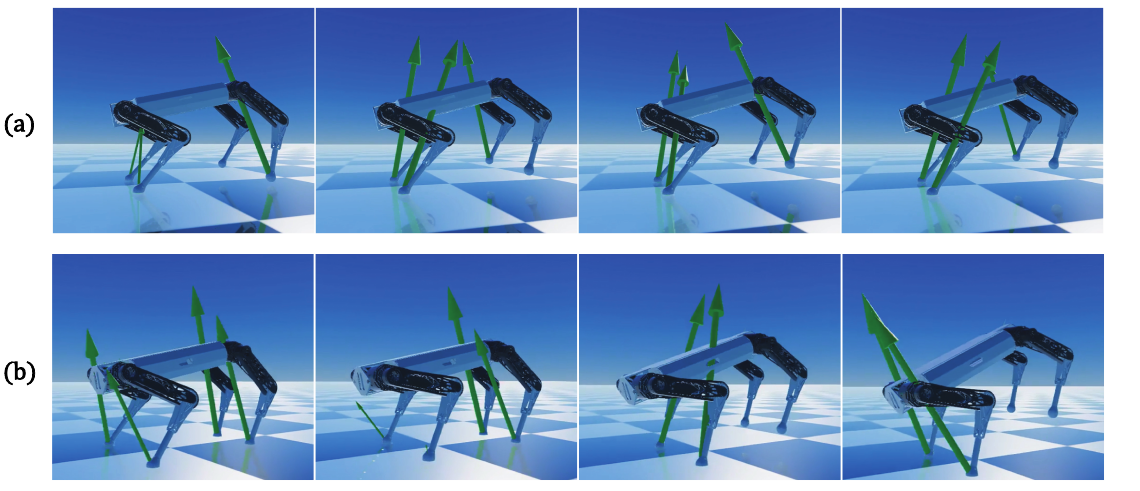}
    \caption[Result of drift compensation]{ Snapshots of the front-leg rearing motion: (a) without and (b) with drift compensation. These depict the solution trajectory with the green arrow showing the contact impulse. In (a), the contact-implicit MPC employs force at a distance, violating the Signorini condition in positional space.}
    \label{figure:resultDrift}
\end{figure}

The adoption of a relatively large discretization time step ($dt=25$ ms) often leads to the occurrence of penetration or `force in the distance' phenomenon. This behavior stems from satisfying the Signorini condition in velocity space, which ensures $v^{n}_{k,i+1}=0$, rather than $\phi_{k,i+1}=0$ in cases of contact. While the penetration phenomenon is less apparent with smaller time steps, such as 1 ms, reducing $dt$ in trajectory optimization is limited by the need to extend the prediction horizon for consistent prediction time. To address this issue, the drift term at the position level~\citep{Yuval2012,carius2019trajectory}, $\frac{\phi_{k,i}}{dt}$, is employed. This term ensures that the approximation of the next time step's foot height approaches a zero value ($v^{n}_{k,i+1}+\frac{\phi_{k,i}}{dt}=0$), as derived in~\eqref{eq:drift}. By integrating the drift term, violations of the Signorini condition in positional space (e.g., force exerted at a distance) are mitigated, as demonstrated in Figure~\ref{figure:resultDrift}.

While the drift compensation for contact-implicit trajectory optimization is already introduced in~\citep{carius2019trajectory}, our study further includes its effect in the gradient. By analytically computing the gradient of contact impulse, we seamlessly incorporate the gradient of the drift component (which reduces to the contact Jacobian).

\subsection*{Gradient Computation for Backward Pass}
We initially compute the analytic gradient of the contact impulse based on the relaxed complementarity constraint. The computed gradients of contact impulse are used in obtaining the gradient $\mathbf{f}_{\mathbf{x}}$ and  $\mathbf{f}_{\mathbf{u}}$ through the gradient of $\mathbf{q}_{i+1}$ and $\mathbf{\dot{q}}_{i+1}$, which are computed from the gradient of $\ddot{\mathbf{q}}_{i}$ through the semi-implicit Euler integration equation. 

The equation of forward dynamics is written as, 
\begin{align}
    \ddot{\mathbf{q}}_{i} = \mathbf{M}_{i}^{-1}(-\mathbf{h}_{i}+\mathbf{B}\mathbf{u}_i+\mathbf{J}_{i}^T\bm{\lambda}_{i}/dt) \nonumber,
\end{align}
and the derivative of the forward dynamics is written as,
\begin{equation}
\label{eq:graident_detail}
\begin{aligned}
    \frac{\partial \ddot{\mathbf{q}}_i}{\partial \mathbf{q}} = & \frac{\partial \mathbf{M}_{i}^{-1}}{\partial \mathbf{q}}(-\mathbf{h}_{i}+\mathbf{B}\mathbf{u}_i+\mathbf{J}_{i}^T\bm{\lambda}_{i}/dt) 
    - \mathbf{M}_{i}^{-1}\frac{\partial \mathbf{h}_i}{\partial \mathbf{q}} \\
    & + \mathbf{M}_{i}^{-1}\frac{\partial \mathbf{J}_i^T}{\partial \mathbf{q}}\bm{\lambda}_i/dt 
    + \mathbf{M}_{i}^{-1}\mathbf{J}_i^T\frac{\partial \bm{\lambda}_i}{\partial \mathbf{q}}/dt , \\ 
    \frac{\partial \ddot{\mathbf{q}}_i}{\partial \dot{\mathbf{q}}} = & - \mathbf{M}_{i}^{-1}\frac{\partial \mathbf{h}_i}{\partial \dot{\mathbf{q}}}
    + \mathbf{M}_{i}^{-1}\mathbf{J}_i^T\frac{\partial \bm{\lambda}_i}{\partial \dot{\mathbf{q}}}/dt , \\ 
    \frac{\partial \ddot{\mathbf{q}}_i}{\partial \mathbf{u}} = & \mathbf{M}_{i}^{-1}\mathbf{B}
    + \mathbf{M}_{i}^{-1}\mathbf{J}_i^T\frac{\partial \bm{\lambda}_i}{\partial \mathbf{u}}/dt . 
\end{aligned}        
\end{equation}
For efficient computation, we first calculate the analytical derivative of the articulated body algorithm (ABA)~\citep{carpentier2018analytical,Pinocchio}, excluding the derivatives of the contact Jacobian, $\mathbf{M}_{i}^{-1}\frac{\partial \mathbf{J}_i^T}{\partial \mathbf{q}}\bm{\lambda}_i/dt$, and the contact impulse component. The frame of the Jacobian should align with the frame of the contact impulse. Therefore, to compute the term $\mathbf{M}_{i}^{-1}\frac{\partial \mathbf{J}_i^T}{\partial \mathbf{q}}\bm{\lambda}_i/dt$, we utilize the kinematic hessian, $\frac{\partial \mathbf{J}_i}{\partial \mathbf{q}}$, in the contact frame. The contact frame aligns with the $\text{CENTERED}$ frame in \citep{kleff2022derivation}. The $\text{CENTERED}$ frame is centered to the local frame, yet its axes remain oriented consistently with the $\text{WORLD}$ frame at all times. Finally, we incorporate the computed gradient of the contact impulse into the equation~\eqref{eq:graident_detail}. As illustrated in Equation~\ref{eq:relaxation_matrix_form2}, computing the gradient of the contact impulse $\frac{\partial \bm\lambda}{\partial \xi}$ requires terms in $\frac{\partial \ddot{\mathbf{q}}_i}{\partial \xi}$. For these terms, we reuse the computed analytical derivative of the ABA without additional computation.

\subsection*{Time Profiling}
\begin{figure}
    \centering
    \includegraphics[width=0.8\columnwidth]{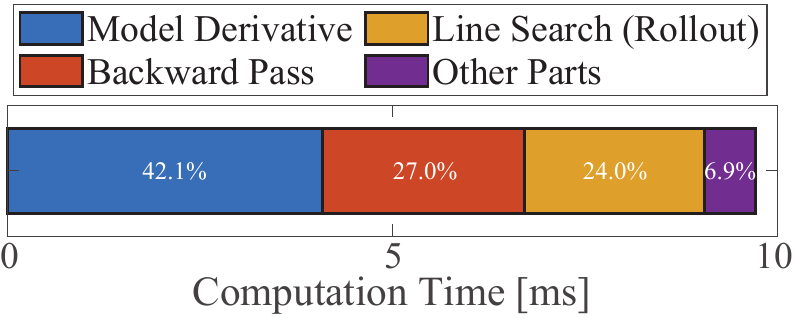}
    \caption[Time profiling for solving an MPC problem.]{Time profiling for solving an MPC problem.}
    \label{figure:time_profiling}
\end{figure}

We conducted time profiling on a desktop PC with an AMD Ryzen 5 3600X processor for an MPC problem with a maximum of 4 iterations to identify computational bottlenecks. Analysis shown in Figure~\ref{figure:time_profiling} reveals that the most demanding task is computing model derivatives, including the gradients of contact impulses and dynamics. The second component is the backward pass, which recursively computes the optimal control step. The third component is the line search process, including the forward roll-out, primarily due to the contact impulse computations.

\subsection*{Weight Setting}
We have created a Table~\ref{tab:weights} to denote the weights for 3D quadruped simulation and experiment, for the regulating cost $l_r$, foot slip and clearance cost $l_f$ with weight $c_f$, air time cost $l_a$ with weight $c_a$, and symmetric control cost $l_s$ with weight $c_s$. The regulating cost weight matrix for state, $\mathbf{W}_{\mathbf{x}}$, is derived by adjusting the positional component, $\mathbf{W}_{\mathbf{q}}$, and the velocity component, $\mathbf{W}_{\dot{\mathbf{q}}}$, with $\mathbf{W}_{\dot{\mathbf{q}}} = \mathbf{W}_{\mathbf{q}} / \alpha$, where $\alpha$ is the velocity weight factor. The final state weight matrix, $\mathbf{W}_{\mathbf{x}_N}$, is calculated as $\beta \mathbf{W}_{\mathbf{x}}$, with $\beta$ representing the final weight factor. The regulating cost weight matrix for control input $\mathbf{W}_{\mathbf{u}}$ is tuned using the diagonal element $c_{\mathbf{u}}$.

\begin{figure*}
    \centering
    \includegraphics[width=2.0\columnwidth]{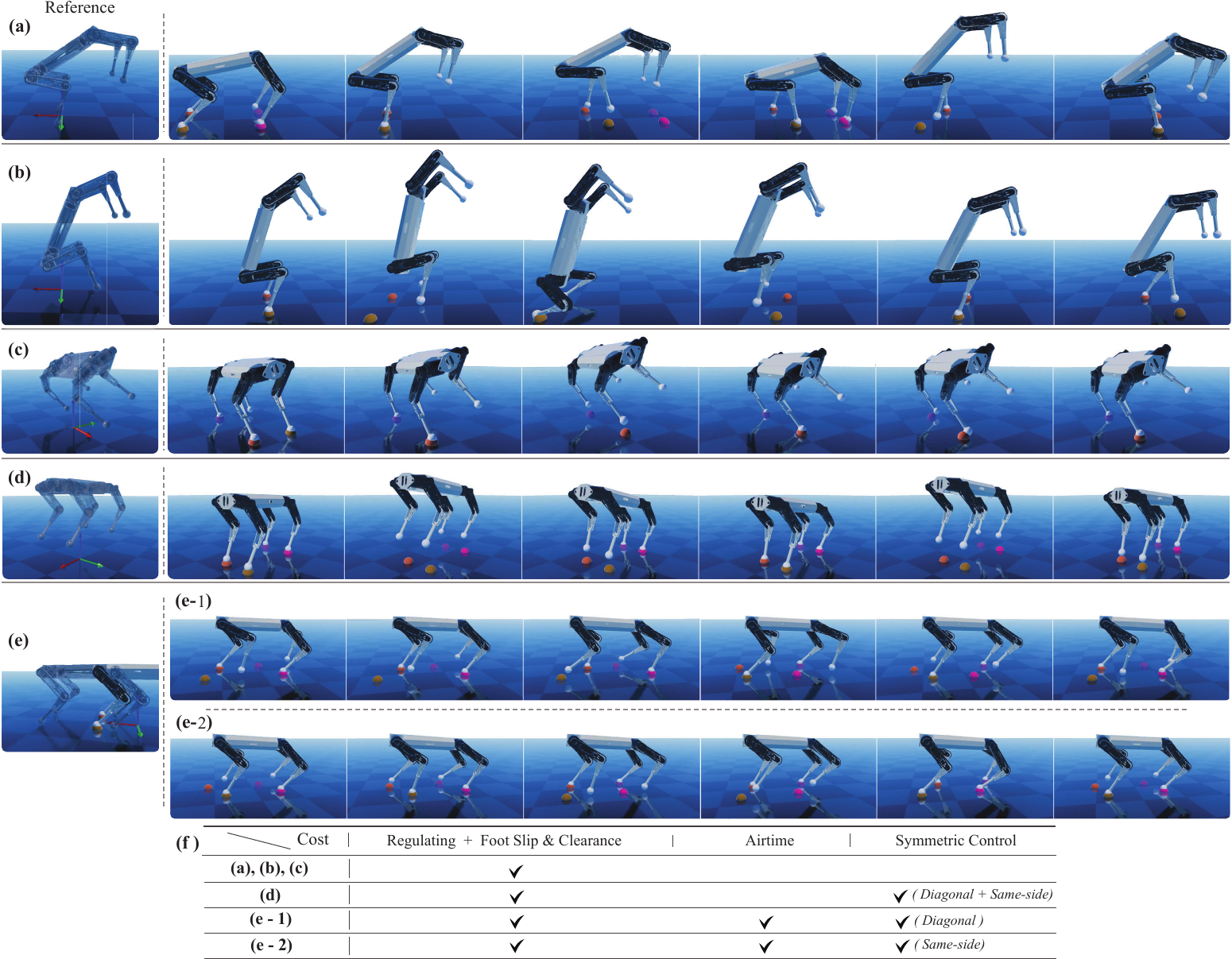}
    \caption[3D simulation result]{3D simulation outcomes for various reference configurations are presented using the proposed contact-implicit MPC approach. The transparent robot on the left side is denoted to the reference configuration, while the robot snapshot depicts the resulting simulation with colored spheres indicating planned next contact points. (a) Target pitch: 0.6 rad (with height offset 0.1 m). (b) Target pitch: 1.2 rad (with height offset 0.1 m). (c) Target roll: 0.6 rad (with height offset 0.1 m). (d) Target height: nominal height + 0.25 m. (e) Target position along the x-axis: current robot x position + 0.55 m. (f) A table detailing cost types for each motion.}
    \label{figure:highlight}
\end{figure*}

For all tasks not related to walking (which includes trotting and pacing), the air time cost weight and symmetric control weight are set to zero: $c_a = 0, c_s = 0$. For fine-tuning in simulations, we change the weights as follows: $c_{\mathbf{u}}=1 \times 10^{-4}$ for the target roll of 0.6 rad, and $\alpha = 1 \times 10^{2}, c_s=1 \times 10^{-3}$ for the target height of +0.25m. In the trotting experiment, we set $\alpha=3.5 \times 10^{1}$ and $c_{\mathbf{u}}=3 \times 10^{-4}$. The snapshots of the resulting motions and the detailed cost types for Extensions 4 and 5 are described in Figure~\ref{figure:highlight}. In the comparison with the relaxed forward model in Section~\ref{sec:comparison_relaxed}, we set the weights for all body parts and joint parts of $l_r$ to $1 \times 10^{1}$ and $1 \times 10^{0}$, respectively, and $\alpha=2 \times 10^{2}$, $c_{\mathbf{u}}=1 \times 10^{-5}$ for both methods. 

\begin{table}[htbp]
\centering
\caption{Default weight setting for 3D quadruped simulation and experiment}
\label{tab:weights}
\begin{tabular}{lll}
\toprule
\textbf{Cost} & \textbf{Parameter} & \textbf{Value} \\
\midrule
$l_r$ & Body position x,y & $2 \times 10^{1}$ \\
& Body position z & $8 \times 10^{1}$ \\
& Body rotation & $1 \times 10^{1}$ \\
& Joint position & $1 \times 10^{0}$ \\
& Control input ($c_{\mathbf{u}}$) & $2 \times 10^{-4}$ \\
\cmidrule{2-3} 
& Velocity weight factor ($\alpha$) & $3 \times 10^{1}$ \\
& Final weight factor ($\beta$) & $1 \times 10^{1}$ \\
\midrule
$l_f$ & Weight ($c_f$) & $1 \times 10^{0}$ \\
\midrule
$l_a$ & Weight ($c_a$) & $2 \times 10^{3}$ \\
\midrule
$l_s$ & Weight ($c_s$) & $1 \times 10^{-2}$ \\
\midrule
\midrule
& Relaxation Variable ($\rho$) & $2 \times 10^{0}$ \\
\bottomrule
\end{tabular}
\end{table}

\section*{Appendix C: Further Discussion}
\label{sec:appendixC}

\begin{figure} [htbp]
    \includegraphics[width=0.8\columnwidth]{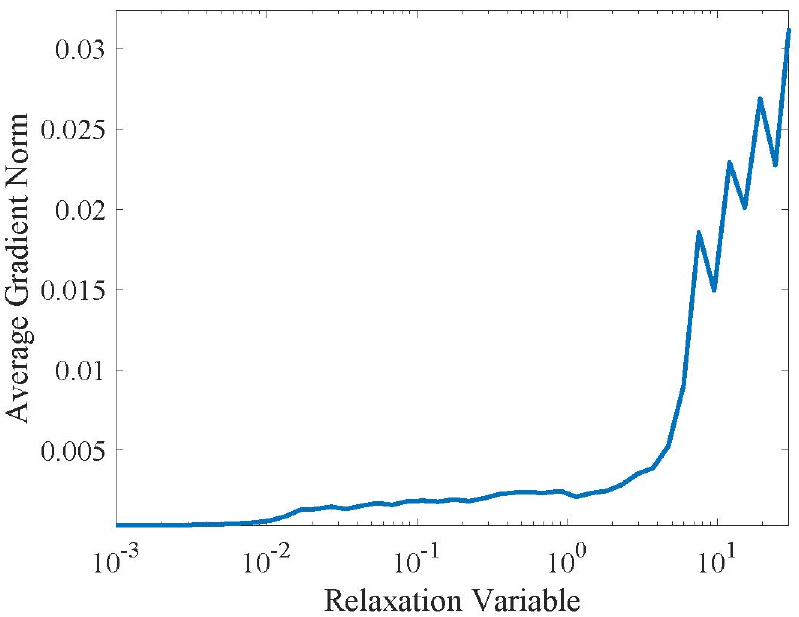}
    \caption[Average gradient norm against relaxation variable]{Average gradient norm at convergence points along relaxation variable for the 2D quadruped jumping task.}
    \label{figure:gradienet}
\end{figure}

\subsection*{Convergence Issue with Excessive Relaxation Values.}
Excessively large relaxation values can cause the relaxed gradient to deviate from the descent direction, potentially degrading convergence. In such cases, the line search process prevents cost increases by halting the step size. Figure~\ref{figure:gradienet} compares the norm of the exact gradient at convergence points for each relaxation variable within the same 2D simulation jumping task as in Figure~\ref{figure:relaxationVariableCost}. The norm of the exact gradient (y-axis of Figure~\ref{figure:gradienet}) indicates how far the convergence point is from the local minima. 

As the relaxation variable increases, the gradient norm progressively rises, positioning convergence points further from the local minima and potentially impeding convergence due to deviations from the descent direction. Notably, there are two abrupt jumps in the gradient norm (Figure~\ref{figure:gradienet}), corresponding with significant cost changes (Figure~\ref{figure:relaxationVariableCost}). When the relaxation variable is very small (less than 0.02), it fails to find the jumping motion. As the relaxation variable increases, it discovers jumping motions, leading to better optimal points (reduced cost) with only a slight increase in the gradient norm; this range of the relaxation is what we intend to utilize. However, values exceeding 2.0 significantly increase both cost and gradient norm; convergence to local minima is not achieved, indicating that the line search fails to reduce the cost.


\subsection*{Dynamics Gap of Front-Leg Rearing Motion.}
\begin{figure}
    \includegraphics[width=1.0\columnwidth]{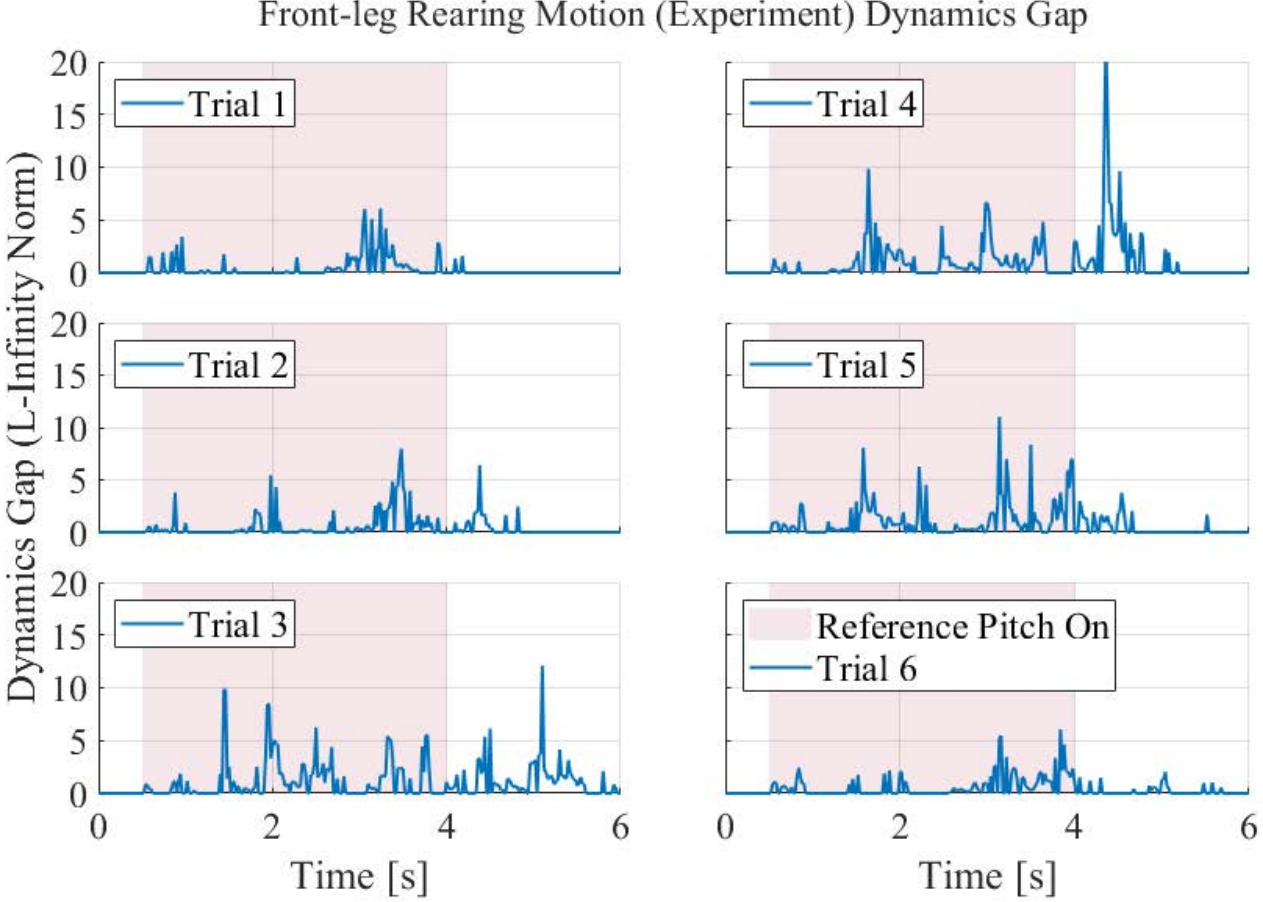}
    \caption[Dynamics gap (after 4 iterations) over time in front-leg rearing experiment]{Dynamics gap (after 4 iterations) over time in front-leg rearing experiment.}
    \label{figure:gap_rearing}
\end{figure}

Figure~\ref{figure:gap_rearing} shows the dynamics gap over time for the front-leg rearing experiment in Section~\ref{result_exp_rearing} for all 6 trials. This figure illustrates instances where the motion is executed without completely closing all dynamics gaps. This is primarily due to the difficulty of the tasks.

Front-leg rearing involves highly dynamic motions with only one or two feet making brief contacts, each critical for the continuation of the motion. Rapid changes in contact and the need for large ground reaction forces to support the robot’s body weight mean that slight contact mismatches can lead to significant deviations, particularly in joint velocity. Despite these challenges, the subsequent MPC iterations close the gaps over time and stabilize the robot. 

In trial 4, which displays the most significant peak value, the robot fails to make contact with the right front leg when landing from rearing poses. In this scenario, feedback PD control initially addresses this issue, followed by a complete motion re-plan in the subsequent MPC problem.

\end{document}